\documentclass{neus2025}

\usepackage{times}
\usepackage{xcolor,colortbl}
\usepackage{caption}
\usepackage{enumitem}
\usepackage{amsmath}
\usepackage{amssymb}
\usepackage{xspace}
\usepackage{listings}
\usepackage{courier}
\usepackage{booktabs}

\usepackage{tcolorbox}
\tcbset{
  colback=white,
  colframe=black,
  boxrule=0.5pt,
  arc=2mm,
  boxsep=3pt,
  fonttitle=\bfseries,
  width=\linewidth,
  before=\vspace{0.2cm},
  after=\vspace{0.2cm}
}
\usepackage{subcaption}

\newcommand{\PySIRTEM}{\texttt{PySIRTEM}\xspace}

\newcommand{\epiExplain}{\texttt{ruleXplain}\xspace}

\title[\epiExplain: Causality by Abstraction]{Causality by Abstraction: Symbolic Rule Learning in Multivariate Timeseries with Large Language Models}
\author{%
 \Name{Preetom Biswas} \Email{pbiswa11@asu.edu}\\
 \addr Arizona State University, Tempe, AZ, USA
 \AND
 \Name{Giulia Pedrielli} \Email{gpedriel@asu.edu}\\
 \addr Arizona State University, Tempe, AZ, USA%
 \AND
 \Name{K. Sel\c{c}uk Candan} \Email{candan@asu.edu}\\
 \addr Arizona State University, Tempe, AZ, USA%
}

\begin{document}

\maketitle

\begin{abstract}
  
Inferring causal relations in timeseries data with delayed effects is a fundamental challenge, especially when the underlying system exhibits complex dynamics that cannot be captured by simple functional mappings. Traditional approaches often fail to produce generalized and interpretable explanations, as multiple distinct input trajectories may yield nearly indistinguishable outputs. In this work, we present \epiExplain, a framework that leverages Large Language Models (LLMs) to extract formal explanations for input-output relations in simulation-driven dynamical systems. Our method introduces a constrained symbolic rule language with temporal operators and delay semantics, enabling LLMs to generate verifiable causal rules through structured prompting. \epiExplain relies on the availability of a principled model (e.g., a simulator) that maps multivariate input time series to output time series. Within \epiExplain, the simulator is used to generate diverse counterfactual input trajectories that yield the same or closely matching target output, serving as candidate explanations. Such counterfactual inputs are clustered and provided as context to the LLM, which is tasked with the generation of symbolic rules encoding the joint temporal trends responsible for the patterns observable in the output times series. A closed-loop refinement process ensures rule consistency and semantic validity. We validate the framework using the \PySIRTEM epidemic simulator, mapping testing rate inputs to daily infection counts; and the EnergyPlus building energy simulator, observing temperature and solar irradiance inputs to electricity needs. For validation, we perform three classes of experiments: (1) we verify the efficacy of the ruleset through input reconstruction; (2) we conduct ablation studies evaluating the causal encoding of the ruleset; and (3) we perform generalization tests of the extracted rules across unseen output trajectories with varying phase dynamics. The code is available \href{https://github.com/Preetom1905011/CausalSeriesExplain}{here}.

\end{abstract}


\section{Introduction and Related Literature}
%\vspace{-0.5em}

Many scientific and engineering domains rely on high-fidelity physics-based simulators that model complex dynamical systems such as epidemic spread, climate projections, and autonomous driving testing. These simulators can accurately generate output timeseries from given inputs, yet they are typically treated as black boxes due to the inner complexity. While they reproduce dynamics, they do not provide transparent explanations of how qualitative behaviors in the output arise from patterns in the input. Timeseries data exacerbate such challenges  due to temporal dependencies, delayed effects, high-dimensionality, and feedback loops, further complicating the characterization of causal dependency. Approaches such as Granger causality~\citep{granger1969investigating}, conditional independence methods~\citep{chu2008search, runge2020discovering}, and Structural Equation Models~\citep{peters2013causal} identify statistical dependencies over the complete simulation horizon.  While insightful, these methods typically quantify aggregate influence rather than providing explicit relations between specific input trends and delayed qualitative output behaviors. These challenges are further complicated when both inputs and outputs are multivariate timeseries produced by a simulator whose structural equations are inaccessible (or intractable).

More recently, Large Language Models (LLMs) have gained prominence in this field due to their pattern recognition, explainable abstraction, and high reasoning capabilities. The attention mechanism in transformers-based LLMs can capture long-range dependencies and complex sequential patterns present in timeseries data~\citep{Time_LLM_sota, TTM2025}. LLM architectures have accelerated timeseries forecasting by managing larger, more complex datasets and capturing non-linearities and dependencies directly from historical data~\citep{liu2022pyraformer, LiuLSTM, liu2024spatialtemporallargelanguagemodel}. For example,~\citet{newsForecastWang2025} combined numerical timeseries and text-based news inputs with chain-of-thought reasoning to connect events with temporal shifts. Also relevant, ~\citep{jin2024timellmtimeseriesforecasting} introduce a framework that reprograms LLMs for timeseries forecasting by mapping timeseries into text-based representations, allowing few-shot and zero-shot prompting without modifying the pre-trained model. Other contributions have explored the construction of causal graphs to predict multi-variate timeseries progression~\citep{orang2025causalgraphfuzzyllms}. Despite such significant strides, few studies have proposed methodologies that attempt identifying generalized, explainable causal relations, especially between multi-variate timeseries inputs and outputs.

In this paper, we propose \epiExplain, a framework that leverages LLMs to infer generalizable, symbolic causal relations between multivariate input and output timeseries with delayed effects. Inspired by in-context learning (ICL)~\citep{ICL}, our approach avoids fine-tuning by providing the LLM with a relevant set of input trajectories, i.e., diverse counterfactual inputs that yield similar output behaviors in a simulator. The LLM is prompted to synthesize a formal ruleset using a constrained symbolic language that captures conjunctions of input trends over time windows and their delayed implications on output phases (e.g., peak onset or plateau formation). These rules are iteratively refined in a closed-loop process: the LLM generates new input configurations based on the extracted rules, the simulator evaluates the resulting outputs, and the ruleset is adjusted to correct inconsistencies or refine the temporal window  associated with each phase. The LLM is instructed to provide symbolic justifications for its decisions, akin to chain-of-thought prompting, but grounded in a formal logical schema. Finally, we evaluate the inference and generalizability of the extracted rules by testing their ability to generate plausible inputs for unseen output trajectories with distinct phase dynamics, validating the causal structure encoding capabilities of \epiExplain.
% Successful reconstruction demonstrates that the rules capture structural properties of the dynamical system rather than memorizing specific trajectories.

To demonstrate the framework, we use timeseries generated from the epidemiological simulator \PySIRTEM~\citep{biswas2025pysirtem}, studying the relationship between testing rates and daily infection trends, and the EnergyPlus simulator~\citep{crawley2001energyplus}, inferring causation between temperature and solar irradiance inputs and electricity consumption for a Supermarket System. Importantly, \epiExplain only uses raw numerical input-output timeseries without any specific semantics associated with epidemics, governing domain-specific dynamics, or underlying system information.

%\vspace{-0.4em}
\paragraph{Contributions.}
%\vspace{-0.5em}
\epiExplain introduces a symbolic and phase-aware framework for explainable causal inference in black-box timeseries systems. Specifically:
%\vspace{-0.5em}
\begin{itemize}\setlength{\itemsep}{2pt}\setlength{\parskip}{0pt}\setlength{\topsep}{1pt}
    \item[(C1)] \textbf{Symbolic framing of delayed causal inference:} We formalize timeseries causal inference as a symbolic rule identification task, introducing a constrained rule language with temporal operators and delay semantics. This shifts the focus from traditional forecasting or graph reconstruction toward interpretable, phase-based reasoning.
    \item[(C2)] \textbf{Domain-agnostic extraction from black-box simulators:} Our method infers interpretable causal abstractions without access to simulator equations or domain priors. It treats multivariate input-output timeseries purely as abstract signals, enabling general-purpose causal rule extraction across black-box systems.
    \item[(C3)] \textbf{LLM-in-the-loop rule synthesis:} We propose a novel prompting and refinement strategy that uses LLMs to generate, verify, and iteratively improve formal rulesets using a closed-loop process grounded in simulation feedback.
    \item[(C4)] \textbf{Generalization via input reconstruction:} We validate inference ability by reconstructing inputs for unseen output trajectories with distinct qualitative dynamics.
\end{itemize}

%\vspace{-0.75em}

%\vspace{-0.9em}
\section{Methodology}
\label{sec:methodology}
\begin{figure}[h]
    \centering
    \includegraphics[width=0.9\linewidth]{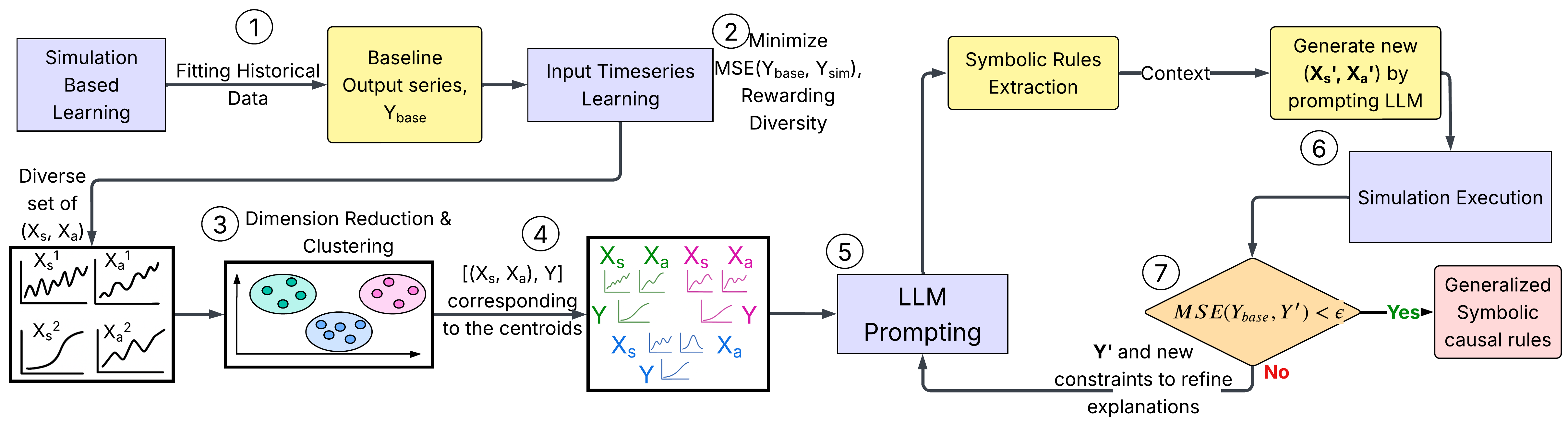}
    \caption{\epiExplain method overview key modules and interactions.}
    \label{fig:model_overview}
\end{figure}

Fig.~\ref{fig:model_overview} summarizes the \epiExplain pipeline.
 
%Starting from a timeseries simulator (e.g., \PySIRTEM)(Sec.~\ref{sec::pySsim}), 
Executing the simulation for a given input, we obtain a baseline output trajectory $Y_{\mbox{\tiny{base}}}$. A learning module then generates diverse multivariate input timeseries that fit $Y_{\mbox{\tiny{base}}}$. These diverse input-output pairs form the basis for extracting causal structure (Sec.~\ref{sec::TRlearn}).

To reduce redundancy, we apply dimensionality reduction and clustering to select representative input-output trajectories (Sec.~\ref{sec::dredClust}). The resulting cluster centroids are used as context for the LLM, which generates a symbolic ruleset $\mathcal{R}$ capturing temporal input trends and their delayed effects on output phases (Sec.~\ref{sec:LLM}).

Next, the LLM generates new input timeseries consistent with $\mathcal{R}$, which are evaluated through the simulator to produce an output $Y'$. The validity of $\mathcal{R}$ is assessed by comparing $Y'$ with $Y_{\mbox{\tiny{base}}}$. If the deviation exceeds a threshold $\epsilon$, the mismatch is used as feedback to refine $\mathcal{R}$ in a closed-loop manner. This iterative process enforces the logical constraints until the ruleset $\mathcal{R}$ yields inputs that produce output trajectories within acceptable deviation from $Y_{\mbox{\tiny{base}}}$ (Sec.~\ref{sec::ruleref}).

%\vspace{-0.5em}
%\vspace{-0.4em}
\subsection{Input Timeseries Learning Module}\label{sec::TRlearn}
%\vspace{-0.4em}

Given a \textit{Baseline Output} trajectory $Y_{\mbox{\tiny{base}}}$ of length $T$, the goal of the learning module is to construct diverse inputs (as multivariate time series) $\mathbf{X}_t$ that would result in similar output $Y_{\mbox{\tiny{sim}}} \approx Y_{\mbox{\tiny{base}}}$. % under the simulator. As mentioned in Sec.~\ref{sec::pySsim}, the input trajectories vary weekly as stepwise constant functions.

We sequentially generate a set of $N$ diverse input time series, $\mathcal{A}_{N} = \left\lbrace \mathbf{X}^p_t \right\rbrace_{p=1}^{N}$
that minimize deviation from the baseline output while maintaining diversity among previously discovered solutions. Here, each component of the input is a time series, i.e., $X_j = \left\lbrace x_{j,1}, x_{j,2}, \ldots, x_{j,T}\right\rbrace$ for each component $j=1,\ldots, J$.%and $X_a = \left\lbrace x_{a,1}, x_{a,2}, \ldots, x_{a,T}\right\rbrace$ where $T$ is the total number of simulated days.
 For each iteration $p=1,\ldots,N$, we solve:
%\vspace{-0.5em}
\begin{align}
P_{\mbox{\tiny{LEARN}}}(p): \min_{\mathbf{X}^{p}} \quad 
& L_{\mbox{\tiny{out}}}(Y_{\mbox{\tiny{sim}}}(\mathbf{X}^{p}), Y_{\mbox{\tiny{base}}}) 
- \lambda_{\mbox{\tiny{arch}}} D_{\mbox{\tiny{arch}}}(\mathbf{X}^{p},\mathcal{A}_{p-1})
\label{eq:TPEmin_prob} \\
\text{s.t.} \quad 
& \ell_{j,t} \le x^{p}_{j,t} \le u_{j,t}, \quad \forall j=1\ldots,J\;\mbox{and}\; t \in \{1,\dots,T\}. \nonumber 
%\\
%& x^{p}_{a,t}, x^{p}_{s,t} \text{ are daily constant.} \nonumber
\end{align}

In problem $P_{\mbox{\tiny{LEARN}}}(p)$, we have:
\begin{align*}
L_{\mbox{\tiny{out}}}(\cdot)
&= \frac{1}{T} \sum_{t=1}^{T} 
\left( Y_{\mbox{\tiny{sim}},t}(\mathbf{X}^{p}_t) - Y_{\mbox{\tiny{base}},t} \right)^2;
\quad
D_{\mbox{\tiny{arch}}}(\cdot)
= \frac{1}{|\mathcal{A}_{p-1}|} 
\sum_{\ell=1}^{|\mathcal{A}_{p-1}|} d^{\ell},
\\[8pt]
\mbox{where: } \quad d^{\ell}
&= \sum_{j=1}^{J}
\left\{
\frac{1}{T} \sum_{t=1}^{T} 
\left| 
\frac{x^{p}_{j,t} - x^{\ell}_{j,t}}{x^{\ell}_{j,t}} 
\right|
\right\}.
\end{align*}

Solving $P_{\mbox{\tiny{LEARN}}}(p)$ sequentially for $p=1,\ldots,N$ yields a set of \textit{diverse} input time series $\left\lbrace (\mathbf{X}^p_t)^T_{t=1} \right\rbrace_{p=1}^{N}$ that produce a similar baseline trajectory $Y_{\mbox{\tiny{base}}}$ (the cost $L_{\mbox{\tiny{out}}}$). To enforce diversity, we maintain an archive $\mathcal{A}$ of previously generated solutions. At iteration $p$, the archive $\mathcal{A}_{p-1}$ contains all time series $\left\lbrace (\mathbf{X}^i_t)^T_{t=1} \right\rbrace_{i=1}^{p-1}$, with $\mathcal{A}_0=\emptyset$, and the newly obtained solution is added after each iteration. The diversity term $D_{\mbox{\tiny{arch}}}$ quantifies the sum of the average difference between the candidate input $(\mathbf{X}^p_t)$ and each archived input across all time steps and input components, weighted by $\lambda_{\mbox{\tiny{arch}}}$ which is chosen by the user to balance diversity.

% %\update{
% Solving sequentially $P_{\mbox{\tiny{LEARN}}}\left(p\right), p=1,\ldots,N$ %is solved $N$ times to generated $N$ 
% produces a set of \textit{diverse} timeseries pairs $\left\lbrace\left(X_s^p, X_a^p\right)\right\rbrace^{N}_{p=1}$ that produce the same output timeseries $Y_{\mbox{\tiny{base}}}$.
% In particular, to ensure the generated solutions are all diverse from one another, we maintain the set $\mathcal{A}$ of solutions that are incrementally generated. As a result, $\mathcal{A}_p$ consists of all the timeseries pairs ($\left\lbrace\left(X^i_s, X^i_a\right)\right\rbrace$) from iterations $i=1,\ldots,p-1$. Equivalently, after each iteration $i$, the obtained solution $(X_s^i, X_a^i)$ is added to $\mathcal{A}_i$. $\mathcal{A}_0$ is initialized as an empty set, i.e., $\mathcal{A}_0=\emptyset$. The deviation, $D_{\mbox{\tiny{arch}}}$, of $(X_s^i, X_a^i)$ from the $\mathcal{A}$ values is determined by the mean absolute percentage error (MAPE) against all values of $\mathcal{A}$. $D_{arch}$ is added as a penalty term with weight $\lambda_{arch}$.
% %}

%\textit{Optimization Method. }
To incrementally solve $P_{\mbox{\tiny{LEARN}}}\left(p\right), p=1,\ldots,N$ we employed the Tree-structured Parzen Estimator (TPE) algorithm~\citep{watanabe2023} in the Optuna environment~\citep{optuna}. 

\subsection{Dimensionality Reduction and Clustering}
\label{sec::dredClust}

To form the training set for the explanation generation step, we apply dimensionality reduction and clustering on the set of timeseries resulting from the learning step (Sec.~\ref{sec::TRlearn}), $\{\left(\mathbf{X}^p_t\right)^{T}_{t=1}\}^{N}_{p=1}$. 

Specifically, we employ the Uniform Manifold Approximation and Projection (UMAP)~\citep{umap2020} technique to represent the data points in a 2-D space embedding and partition into clusters through $K$-Means clustering, from which a single centroid representative $\left(\mathbf{X}^c_t\right)^{T}_{t=1}$ is selected per cluster. The resulting input-output pairs $\{\left(\mathbf{X}^c_t\right)^{T}_{t=1}, Y_{\mbox{\tiny{sim}}}^c)\}$ serve as the training context for the LLM during explanation generation (Sec.~\ref{sec:LLM}). 

This reduction step preserves diversity while eliminating redundant examples that could bias the learned ruleset toward over-specific patterns. Implementation details are provided in Appendix~\ref{app::dredClust}.

\subsection{LLM-based Explanation Generation}
\label{sec:LLM}
The explanation generation component of the proposed \epiExplain consists of two steps: (i) initial ruleset elicitation based on the cluster-derived context and (ii) closed-loop rule refinement process. The specific LLM used in this work is the OpenAI gpt4.1~\citep{openai2024gpt41}. %\update{
The complete procedure is also summarized in Table~\ref{tab:rule_refinement}.%}

\paragraph{Initial Explanation Inference.}\label{sec::expInf}
%Once we have the data points grouped into distinct clusters, 
\ignorespacesafterend
This step receives as input the centroids resulting from the clustering (Sec.~\ref{sec::dredClust}), namely %we select the input-output pairs corresponding to each cluster centroids 
$\left(\left(\mathbf{X}^p_t\right)^{T}_{t=1}, Y_{\mbox{\tiny{sim}}}^c\right)$, which we refer to as the \textit{context}, for generating the initial set of explainable symbolic rules that characterize the relationship between input and output timeseries. Notably, we avoid providing \epiExplain with prior knowledge of the underlying system dynamics (e.g., epidemics-specific transmission dynamics) and domain-specific notations. Instead, %we adopt general functional terminology such as $X$ for the testing rates and $Y$ for the infection cases. The 
variables are presented purely as abstract timeseries data without any explicit reference to their physical meaning and/or regulating equations, conveying only the existence of a causal relationship. Since $Y_{\mbox{\tiny{sim}}}^c \approx Y_{\mbox{\tiny{base}}}$, we will represent the contexts with $\{\left(\mathbf{X}^c_t\right)^{T}_{t=1}, Y_{\mbox{\tiny{base}}})\}$.

Along with the aforementioned context, through prompting, we further guide the LLM to infer a causal ruleset by stating the following general properties: (i) %, the LLM is slightly guided by information about the general direction of the relation, i.e., 
direct/inverse trend between $\left(\mathbf{X}^c_t\right)^{T}_{t=1}$ and $Y$, %but not definitively stating any 
without providing an indication of proportionality, (ii) %We also inform of the presence of a 
delayed effect (temporal lag) between the input (cause) and the output series (effect) and, (iii) explicit constraints on structure, format, and vocabulary to ensure reproducibility and consistency (Sec.~\ref{sec::formalrep}). The specific prompt to generate the initial set of rules is listed in Appendix~\ref{app:supple}.

\paragraph{Closed-loop Rule Refinement.}\label{sec::ruleref}

Given the initial ruleset, the closed-loop refinement pipeline iterates until an \textit{acceptable} set of rules is obtained, i.e., when the input sequence generated from the ruleset yields an output $Y_{\mbox{\tiny{sim}}}$  sufficiently similar to the baseline output. Without loss of generality, we say that two time series are similar when the distance measure $L_{\mbox{\tiny{out}}}(Y_{\mbox{\tiny{sim}}}, Y_{\mbox{\tiny{base}}})$ is bounded by a user selected positive $\epsilon$.
%}
The refinement iterates through three key steps:
\begin{enumerate}[label=\roman*)]
\setlength{\itemsep}{1.5pt}\setlength{\parskip}{0pt}\setlength{\topsep}{1pt}
    \item \textit{\textbf{Input set generation using the initial ruleset}}. Given the cluster-derived context, we prompt the LLM to generate a unique input timeseries $\mathbf{X}' = (\mathbf{x}_{1}', \dots, \mathbf{x}_{T}')$ %and $X_a' = (x_{a,1}', \dots, x_{a,T}')$ 
    that differs from the cluster-derived inputs ($\left\lbrace\left(\mathbf{X}^c_t\right)^{T}_{t=1}\right\rbrace^C_{c=1}$) but remains strictly consistent with the initial ruleset. 
    We prompt the LLM to identify the different phases, predefined by a schema, and generate the input configurations accordingly. Moreover, the LLM is instructed to justify the generated input values to prevent misalignment with the previously inferred relationships.
    \item \textit{\textbf{Evaluation of the new input.}} The generated input timeseries are evaluated by executing the simulator, obtaining the corresponding output trace $Y'$. All other simulator hyperparameters remain unmodified.
    \item \textit{\textbf{Ruleset refinement.}} The new input-output pair $\left\lbrace\left(\mathbf{X}'_t\right)^{T}_{t=1}, Y'\right\rbrace$ is then returned to the LLM model along with the baseline $Y_{\mbox{\tiny{base}}}$ %for deviation assessment, i.e., to quantify how different the trajectories of 
    to quantify the deviation $L_{\mbox{\tiny{out}}}(Y_{\mbox{\tiny{sim}}}, Y_{\mbox{\tiny{base}}})$. The LLM also retains the cluster contexts and prior iteration results. By evaluating the $\left\lbrace\left(\mathbf{X}'_t\right)^{T}_{t=1}, Y'\right\rbrace$ and $Y_{\mbox{\tiny{base}}}$, the LLM identifies inaccurate symbolic rules and generates a \textit{modified} (refined) ruleset. Additional qualitative constraints (such as ``$Y'$ displays weaker resurgence than $Y_{\mbox{\tiny{base}}}$'', and ``the inputs should not oscillate'') are provided as constraints to guide ruleset refinement. The rationale behind iterative refinement with progressive constraints, instead of offering all the constraints initially, is to reduce reasoning load and avoid overwhelming the model. At this stage of \epiExplain development, the constraints are added manually based on observed deviations in $Y_{\mbox{\tiny{base}}}$ and $Y'$. 
\end{enumerate}

\paragraph{Formal Representation of Extracted Rules}\label{sec::formalrep}
To formalize the structure of the extracted rules, we adopt a symbolic and temporal logic-based notation to describe causal dependencies between trends in the input timeseries and phases of the output. Each rule is represented as a temporal implication of the form:
{\small
\begin{equation}
\label{eq:causal_rule}
\Box_{[t_1, t_2]} 
\left( \textsf{Trend}_h(X_h) \land \textsf{Trend}_j(X_j) \right)
\rightarrow 
\Diamond_{[t_3, t_4]} \textsf{Phase}(Y)
\end{equation}
}
where $h$ and $j$ are two particular dimensions of the input timeseries. Here:
\begin{itemize}
    \setlength{\itemsep}{2pt}
    \setlength{\topsep}{1pt}
    \setlength{\parskip}{0pt}
    \item $\textsf{Trend}_\cdot(X_\cdot)$ is a unary predicate applied to an input series over interval $[t_1, t_2]$, drawn from a fixed vocabulary $\mathcal{T}$ (e.g., $\textsf{Low}$, $\textsf{SharpSpike}$, $\textsf{HighStable}$).
    \item $\textsf{Phase}(Y)$ is a qualitative output behavior over $[t_3, t_4]$, drawn from a vocabulary $\mathcal{P}$ (e.g., $\textsf{InitialGrowth}$, $\textsf{PeakFormation}$).
    \item $\Box$ and $\Diamond$ denote persistent and eventual satisfaction of input and output properties over their respective intervals.
    \item Rules may include explicit delay constraints: $t_3 > t_2$.
    \item The delay $\delta = t_3 - t_2$ captures the temporal lag between cause and effect.
\end{itemize}

As an illustrative example, consider the rule:
{\small
\begin{equation}
\label{eq:example_rule}
\Box_{[t_1, t_2]} 
\left( \textsf{SlowRise}(X_h) \land \textsf{Low}(X_j) \right)
\rightarrow 
\Diamond_{[t_3, t_4]} \textsf{PeakFormation}(Y)
\end{equation}
}
% %\vspace{-0.2em}
This rule states that if $X_h$ slowly rises and $X_j$ remains low over a time window $[t_1, t_2]$, then a peak in $Y$ will occur in the delayed interval $[t_3, t_4]$. The ruleset $\mathcal{R}$ is composed of multiple such formulas, one per output phase, and provides an interpretable symbolic explanation of the causal dynamics. 
%These rules enable generation of consistent counterfactual inputs and support iterative refinement under the closed-loop learning process described in Sec.~\ref{sec::expInf}-\ref{sec::ruleref}.

\begin{table}[h]
\centering
\footnotesize
\renewcommand{\arraystretch}{1.2}
\captionsetup{font=small}
\setlength{\tabcolsep}{8pt}
\caption{Stepwise process for Causal Rules Inference through LLM.}
\begin{tabular}{p{0.95\columnwidth}}
\toprule
\textbf{Procedure: Explanation Generation and Refinement} \\
\midrule
\textbf{Step 1: Initialize.} Extract an initial ruleset \(R_0\) derived from Cluster-derived context $[\left(\mathbf{X}^c_t\right)^{T}_{t=1}, Y_{base}]_{c=1}^C$ pairs. \\[0.4em]
\textbf{Step 2: Generate Inputs.} Use \(R_i\) and $[\left(\mathbf{X}^c_t\right)^{T}_{t=1}, Y_{base}]_{c=1}^C$ to prompt novel input sequences \(\left(\mathbf{X}^{(i)}_t\right)^{T}_{t=1})\) that approximately reproduce the baseline trajectory \(Y_{\text{base}}\). \\[0.4em]
\textbf{Step 3: Simulate \& Evaluate.} Run the simulator and compute deviation \(\Delta^{(i)}\) using the loss $L_{\mbox{\tiny{out}}}$ calculated from $Y^{(i)}$ and $Y_{base}$. \\[0.4em]
\textbf{Step 4: Update Rules.} If \(\Delta^{(i)} > \epsilon\), update \(R_i\) based on observed mismatches and added constraints to get \(R_{i+1}\). \\[0.4em]
\textbf{Step 5: Repeat.} Continue Steps 2–4 until \(\Delta^{(i)} \leq \epsilon\) or the maximum number of iterations is reached, yielding the refined ruleset \(R_{\mbox{\tiny{final}}}\). Please note that $[\left(\mathbf{X}^c_t\right)^{T}_{t=1}, Y_{base}]_{c=1}^C$ remain unchanged across iterations. \\
\bottomrule
\end{tabular}
\label{tab:rule_refinement}
\end{table}

\section{Numerical Experiments}\label{sec::results}
%\vspace{-0.5em}
In this section, through set of validation experiments, we address three questions:
\begin{itemize}\setlength{\itemsep}{1pt}\setlength{\parskip}{0pt}\setlength{\topsep}{1pt}
    \item[Q1.] Do the rules generated within \epiExplain lead to  increasingly more accurate and explainable timeseries? %the accuracy and effectiveness of the ruleset $R_{final}$ in regenerating novel input trajectories. 
    \item[Q2.] How much does the ruleset or context impact the quality of the generated input timeseries?
    \item[Q3.] Is the ruleset generalizable to any baseline timeseries?
\end{itemize}
%\update{
% We address Q1 in Sec.~\ref{sec::resQ1} by observing the output deviation (MSE between $Y'$ and $Y_{base}$) across rule refinement iterations. In Sec.~\ref{sec:context_sec}, we investigate Q2 by experimenting with different prompt strategies (e.g. providing only rulesets or the combination of cluster-derived context and ruleset) to gauge the quality of the produced input solutions. Finally, we test Q3 in Sec.~\ref{sec::gen_sec} applying the ruleset to generate inputs for varying, unseen, output timeseries.
% For experimental purposes, we adopt: (i) an epidemic use case, mapping human-regulated testing rates to daily infection counts; and (ii) a building energy simulation task, inferring the relationship between weather variables--temperature and solar irradiance--and electricity consumption.
% % \textcolor{red}{Preetom briefly mention input output and such...}. 
% Sec.~\ref{sec::pySsim} briefly introduces the simulation tool adopted.
% For experimental evaluation, we consider two use cases: (i) an epidemic scenario linking human-regulated testing rates to daily infection counts, and (ii) a building energy simulation task relating weather variables (temperature and solar irradiance) to electricity consumption. Sec.~\ref{sec::pySsim} briefly introduces the adopted simulation tool.
\subsection{Simulation Environments}
\label{sec::pySsim}
%\vspace{-0.5em}
To demonstrate \epiExplain, we use different two simulation environments:
\begin{itemize}[topsep=1pt, itemsep=1pt,parsep=0pt]
    \item \PySIRTEM~\citep{biswas2025pysirtem}. For this epidemic simulator, the relevant input-output pair for the causal analysis is $\{\left(\mathbf{X}^p_t\right)^{T}_{t=1},Y \}\equiv\{ (X_{s,t}, X_{a,t}), Y \}$, where $X_s$ and $X_a$ denote the symptomatic and asymptomatic testing-rate, respectively, and $Y$ denotes the daily infected cases over $T = 100$ days. The input trajectories vary weekly, consistent with the simulator assumptions.
    \item EnergyPlus~\citep{crawley2001energyplus}. For this building energy simulator, the relevant input–output pair is ${(X_d, X_g), Y}$, where $X_d$ denotes Dry Bulb Temperature ($^oC$), $X_g$ denotes solar input, i.e., the global horizontal irradiance ($Wh/m^2$), and $Y$ denotes building electricity usage ($J$) for a supermarket setting. Inputs vary hourly and horizon is $T = 100$ hours. 
\end{itemize}
%first use timeseries generated from the epidemiological simulator  We further evaluate \epiExplain using the 

In our experiments, we treat the simulator as a black box: its internal equations and parameters remain fixed and are not used by \epiExplain. A detailed description of the \PySIRTEM and EnergyPlus model dynamics is provided in Appendix~\ref{app:pysirtem}-~\ref{energyPlus_sim}.

\subsection{Q1: Evaluation of Ruleset-Generated Input Timeseries}\label{sec::resQ1}
%\vspace{-0.5em}
To address Q1, we apply \epiExplain to extract symbolic rules describing the relationship between multivariate input time series and the target output trajectory. For each simulation environment, we first select a baseline output time series $Y_{\text{base}}$ over a fixed horizon ($T=100$). For \PySIRTEM, we used the daily infection data from March 22 to June 30, 2020~\citep{ctw2021} and for EnergyPlus, the electricity consumption for a supermarket setup in Miami, FL from Jan 1 to Jan 5, 1997. From the baseline output, 
% \epiExplain generates a diverse set of candidate input trajectories as context (Sec.~\ref{sec::TRlearn},~\ref{sec::dredClust}) and iteratively infers and refines a symbolic ruleset as described in Sec.~\ref{sec:LLM}.
we solved $P_{\mbox{\tiny{LEARN}}}(p)$ for $p=1,\ldots,N=20$ obtaining  a preliminary set %$\mathcal{S}$, 
$\mathcal{A}$ with %$|\mathcal{S}|=20$ 
$|\mathcal{A}|=20$ top  performing input timeseries $\left\lbrace(X^{\ell}_s, X^{\ell}_a)\right\rbrace^{|\mathcal{A}|}_{\ell=1}$ ranked by distance measure from the baseline, $L_{out}$. The centroid timeseries $\left\lbrace(\mathbf{X}^c_t)^T_{t=1}\right\rbrace^{C}_{c=1}$, obtained from U-MAP projection and K-Means clustering, were used as input to the simulators (\PySIRTEM and EnergyPlus) to produce the  $Y_{\mbox{\tiny{sim}}} \approx Y_{\mbox{\tiny{base}}}$.  %and generate output  
% \begin{figure}[t]
%     \centering
%     \includegraphics[width=0.7\linewidth]{Figs/cluster.png}
%     \caption{Clustering of the initial solution pool $\left\lbrace (X^p_s, X^p_a)\right\rbrace_p$ from solving $P_{\mbox{\tiny{LEARN}}}(p)$.}
%     \label{fig:clusters}
% \end{figure}
Given the clusters (context) $\left\lbrace(\mathbf{X}^c_t)^T_{t=1}, Y_{\mbox{\tiny{base}}}\right\rbrace$, we can then execute the procedure detailed in Sec.~\ref{sec:LLM}, where we infer the formal rulesets $\mathcal{R}_{\mbox{\tiny{final}}}$ for both \PySIRTEM and EnergyPlus (complete rulesets are reported in Appendix~\ref{app:causal_ruleset_sec}, Table~\ref{tab:causal_rules}-~\ref{tab:causal_rules_eng}). %\update{

\begin{figure}[t]
  \centering

  % first row
  \begin{minipage}[b]{0.5\textwidth}
    \centering
    \includegraphics[width=\linewidth]{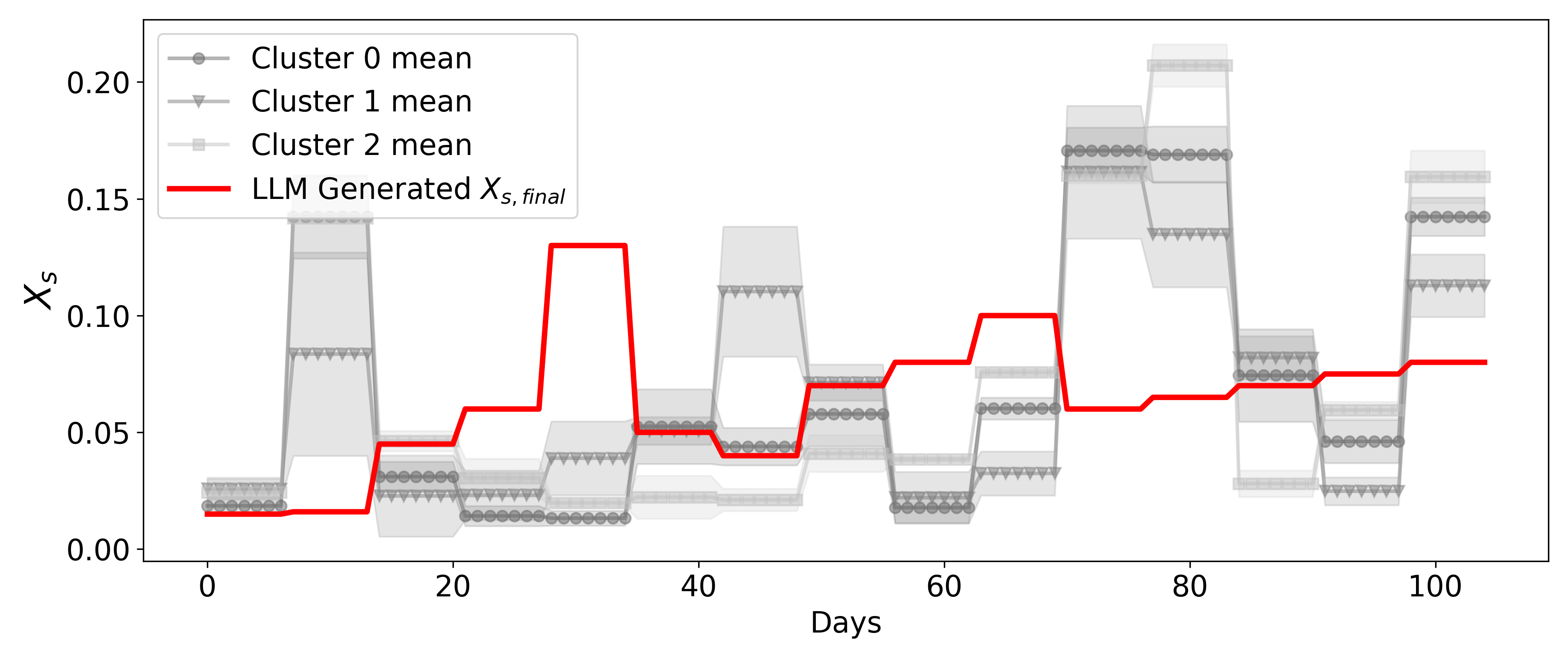}
  \end{minipage}\hfill
  \begin{minipage}[b]{0.5\textwidth}
    \centering
    \includegraphics[width=\linewidth]{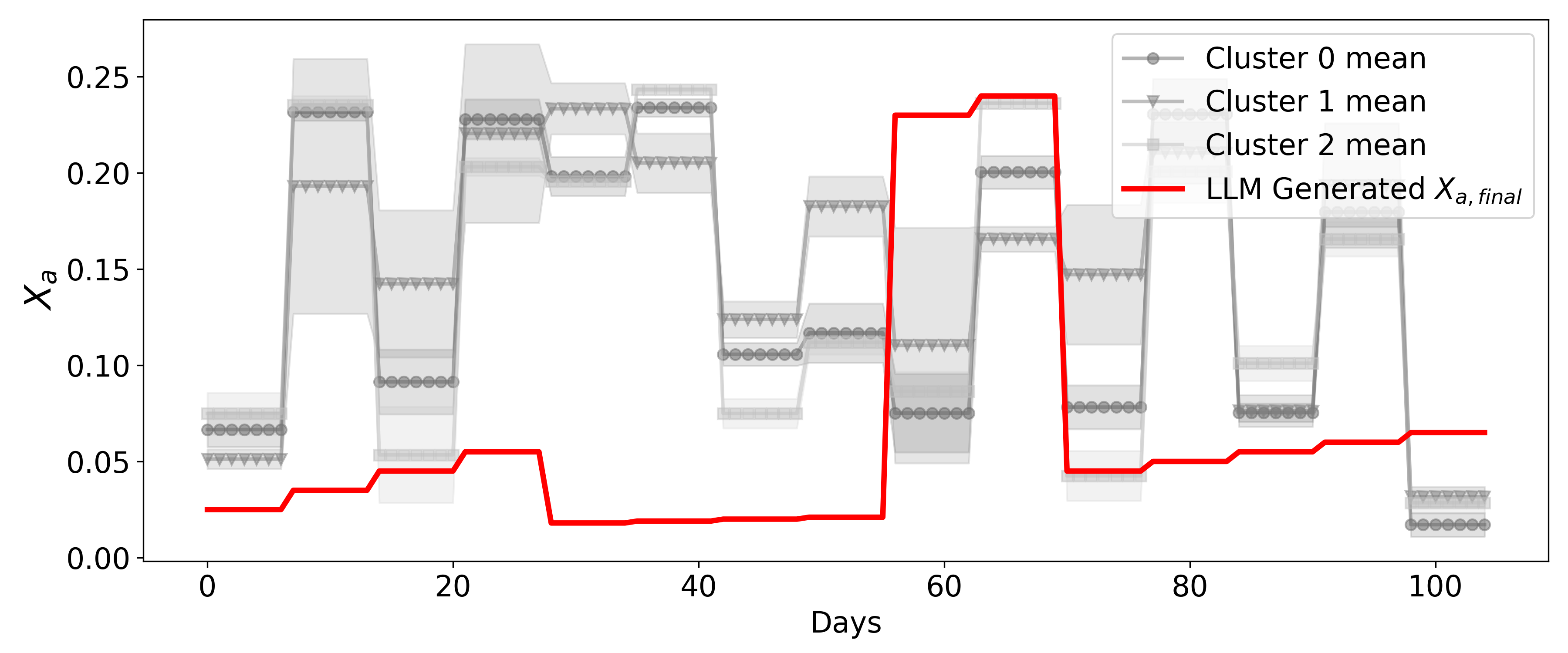}
  \end{minipage}

  % \vspace{0.5cm}

  % second row
  \begin{minipage}[b]{0.5\textwidth}
    \centering
    \includegraphics[width=\linewidth]{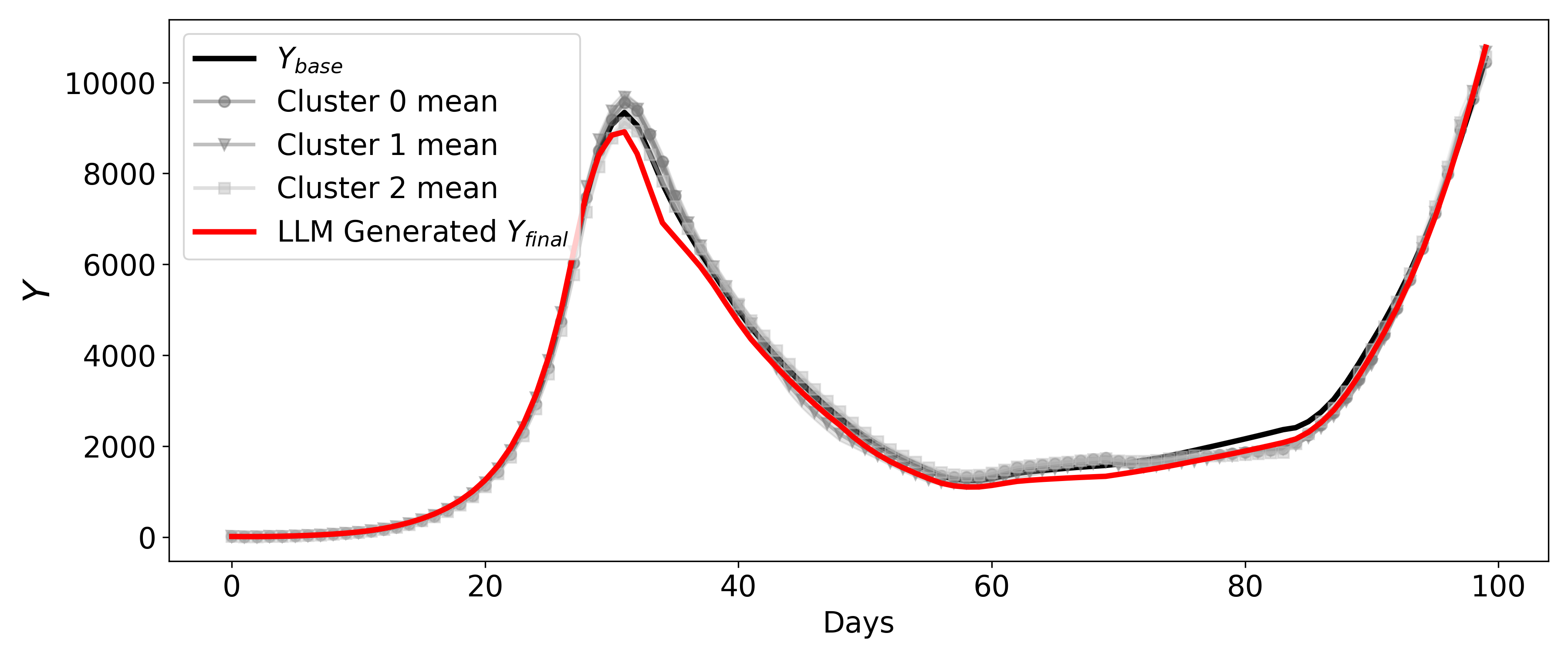}
  \end{minipage}

  \caption{Timeseries $X_s$, $X_a$, and $Y$ generated using $\mathcal{R}_{final}$ for \PySIRTEM.}
  \label{fig:cluster_time_pysim}
\end{figure}

\begin{figure}[t]
  \centering

  % first row
  \begin{minipage}[b]{0.5\textwidth}
    \centering
    \includegraphics[width=\linewidth]{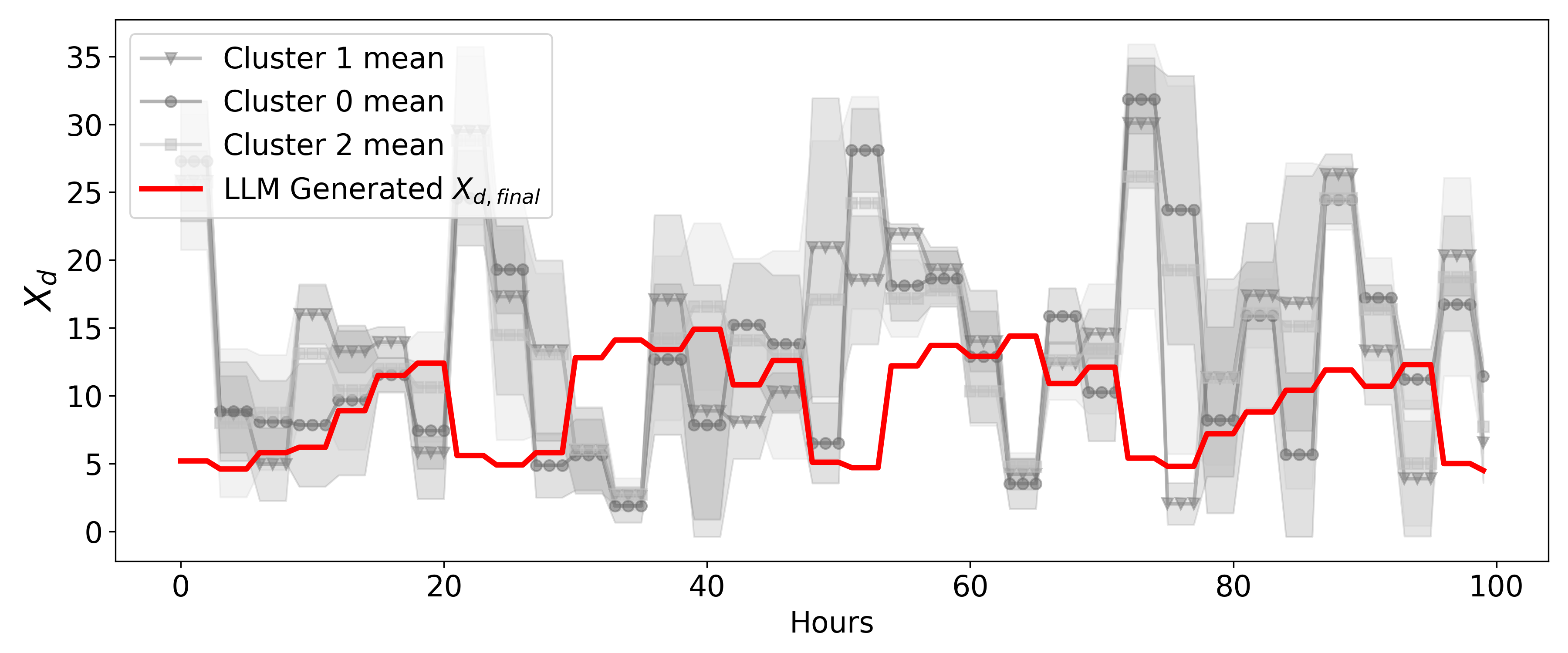}
  \end{minipage}\hfill
  \begin{minipage}[b]{0.5\textwidth}
    \centering
    \includegraphics[width=\linewidth]{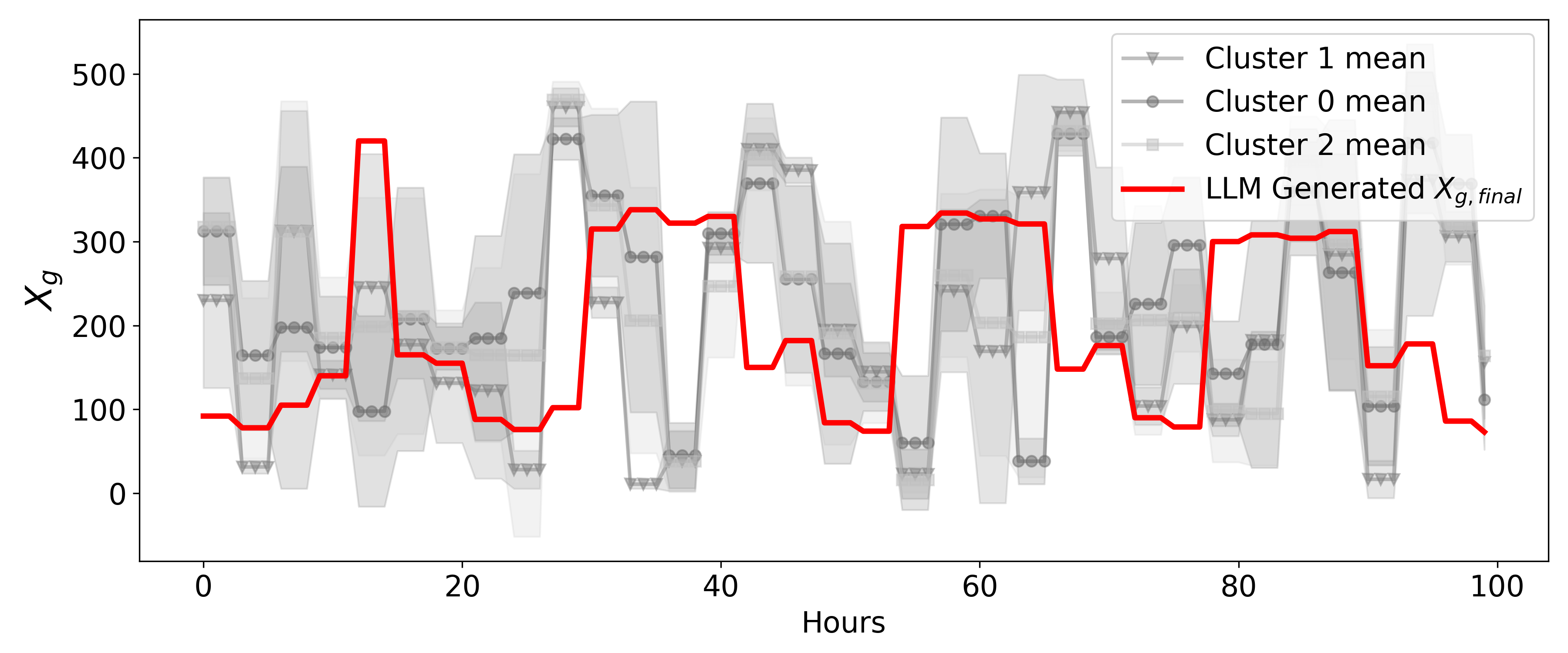}
  \end{minipage}

  % \vspace{0.5cm}

  % second row
  \begin{minipage}[b]{0.5\textwidth}
    \centering
    \includegraphics[width=\linewidth]{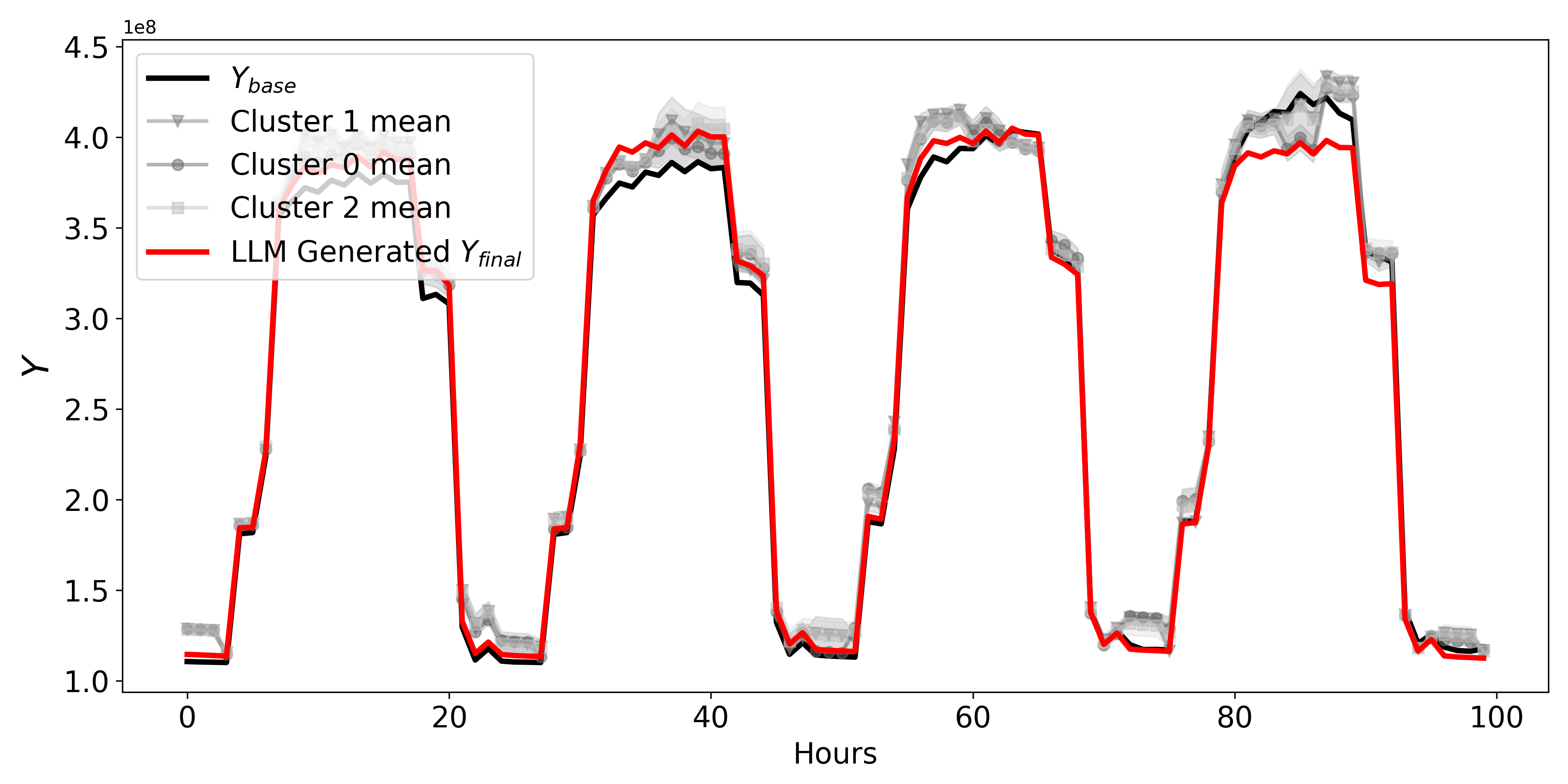}
  \end{minipage}

  \caption{Timeseries $X_d$, $X_g$, and $Y$ generated using $\mathcal{R}_{final}$ for EnergyPlus.}
  \label{fig:cluster_time_engplus}
\end{figure}

Fig.~\ref{fig:cluster_time_pysim}-~\ref{fig:cluster_time_engplus} shows the input–output trajectories resulting from the final ruleset $\mathcal{R}_{\mbox{\tiny{final}}}$ produced by \epiExplain, for both \PySIRTEM and EnergyPlus respectively, along with centroid context timeseries. The close temporal and magnitude alignment of the simulated $Y_{\mbox{\tiny{final}}}$ and $Y_{\mbox{\tiny{base}}}$ is indicative of the accuracy and effectiveness of $\mathcal{R}_{\mbox{\tiny{final}}}$ in explaining the inherent (and unknown) dynamics emerging from the principled simulation. In particular, for EnergyPlus, the periodic structure of the output signal is also effectively reflected through the inferred input patterns. Across all our experiments we consistently observed lower errors as a function of the \epiExplain iterates, thus supporting convergence of the proposed approach (numerical evidence is reported in Appendix~\ref{app:causal_ruleset_sec}, Fig.~\ref{fig:RQ1}).

%in the Appendix illustrate the downward learning difference measure trend across iteration, solidifying the efficacy of the iterative refinement process in learning.

% \begin{figure}[htbp]
%     \centering
%         \includegraphics[width=0.75\linewidth]{Figs/cluster_timeseries.png}
%         \caption{Timeseries $X_s$, $X_a$, and $Y$ generated using $\mathcal{R}_{final}$.}
%         \label{fig:cluster_time}
% \end{figure}

% \begin{figure}[htbp]
%   \centering

%   % first row
%   \begin{minipage}[b]{0.46\textwidth}
%     \centering
%     \includegraphics[width=\linewidth]{Figs/cluster_timeseries_pysim.png}
%   \end{minipage}
%   \begin{minipage}[b]{0.46\textwidth}
%     \centering
%     \includegraphics[width=\linewidth]{Figs/cluster_timeseries_eng.png}
%   \end{minipage}
%   \caption{(a) Timeseries $X_s$, $X_a$, and $Y$ generated using $\mathcal{R}_{final}$ for \PySIRTEM (b) Timeseries $X_d$, $X_g$, and $Y$ generated using $\mathcal{R}_{final}$ for EnergyPlus.}
%   \label{fig:cluster_time}
% \end{figure}

%\vspace{-0.75em}
\subsection{Q2: Impact of Ruleset and Cluster-Derived Context}
\label{sec:context_sec}

We examine how much causal information is encoded in the causal ruleset and the cluster-derived context by using them to generate input timeseries corresponding to a fixed target output $Y_{\mbox{\tiny{base}}}$ in our \PySIRTEM experiment. Specifically, we evaluate three scenarios: \textbf{(S1)} \textit{no ruleset and no clustering context}, \textbf{(S2)} \textit{ruleset only}, and \textbf{(S3)} \textit{ruleset with clustering context}. This comparative setup allows us to isolate and quantify the contribution of $\mathcal{R}_{\mbox{\tiny{final}}}$ and contextual examples to the accuracy and novelty of the generated inputs. The specific prompts for each experiment are listed in the Appendix~\ref{app:val_prompts}. 

\begin{figure*}[htbp]
    \centering
    \includegraphics[width=0.85\linewidth]{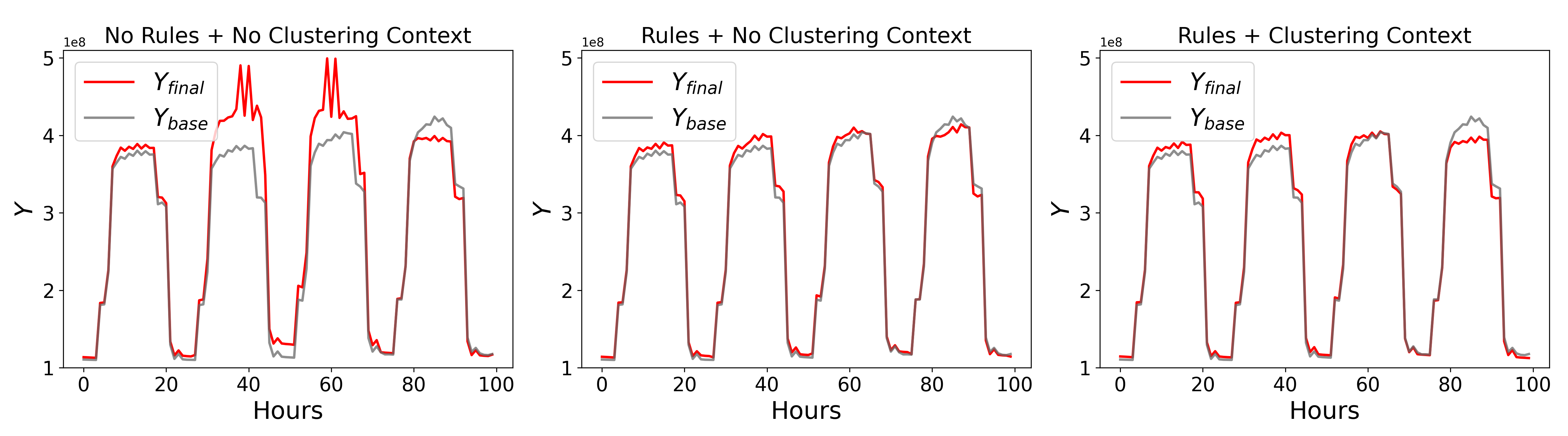}
    \caption{Causal Encoding Comparison of $\mathcal{R}_{\mbox{\tiny{final}}}$ through scenarios S1-S3 for EnergyPlus.}
    \label{fig:context_figs_engplus}
\end{figure*}

Fig.~\ref{fig:context_figs_pysirtem}-~\ref{fig:context_figs_engplus} display a clear progressive improvement from S1 to S2. The no-guidance (S1) scenario produces outputs with the least accuracy, which is significantly increased with $\mathcal{R}_{\mbox{\tiny{final}}}$ (S2). Adding clustering context (S3) yields mixed effects, reporting slight improvement for \PySIRTEM, but less flexible to baseline trends for EnergyPlus. This suggests potential overfitting to the examples rather than improved adherence to the baseline and ruleset. The progression, particularly the major gain from S1 to S2, corroborate the encoding ability of causal information in the \epiExplain-generated ruleset, while addition of context in downstream tasks could overbias the model.

% Even with only $\mathcal{R}_{\mbox{\tiny{final}}}$ (Fig.~\ref{fig:context_figs}, center column), the generated input remains consistent with the ruleset and the output error remains controlled, except for a slight increase in the final segment, likely due to accumulation of small simulation errors.

Across all scenarios, the timing of key phases (e.g. peak and decline) are aligned with $Y_{\mbox{\tiny{base}}}$. We conjecture it to be due to the underlying physics of system: specific parameters govern certain qualitative properties. For example, in \PySIRTEM, infection rate is the primary driver of the disease progression which is kept constant to infer causal dependency between decidable human-regulated testing rates and corresponding infection case trends. Similarly, in EnergyPlus, we isolate the influence of observable weather inputs (e.g., temperature) on facility-level electricity consumption. In both cases, our focus was to generate interpretable insights that can support decision makers.
%}

\begin{figure*}[htbp]
    \centering
    \includegraphics[width=0.9\linewidth]{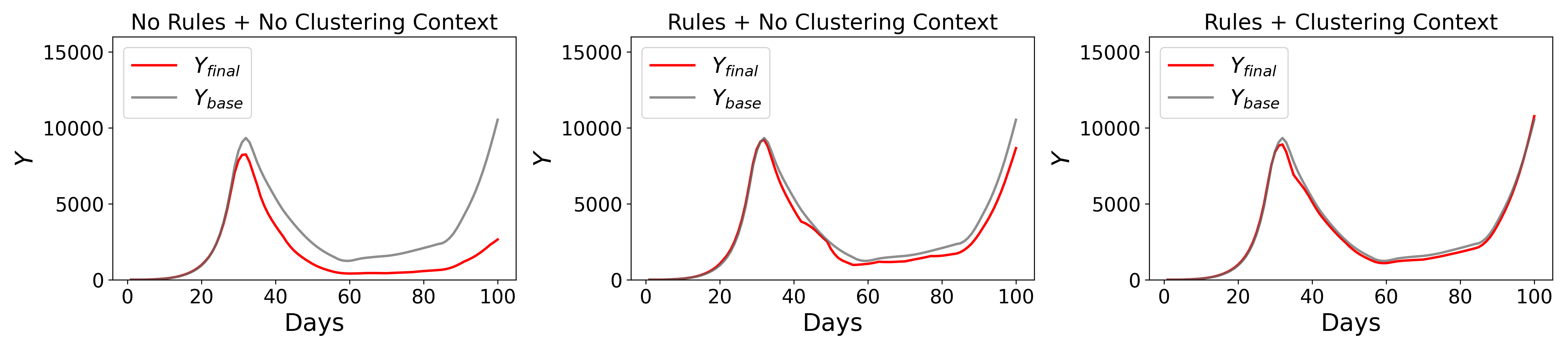}
    \caption{Causal Encoding Comparison of $\mathcal{R}_{\mbox{\tiny{final}}}$ through scenarios S1-S3 for \PySIRTEM.}
    \label{fig:context_figs_pysirtem}
\end{figure*}

%\vspace{-0.75em}
\subsection{Q3: Results Generalizability Across Baselines}
\label{sec::gen_sec}
%\vspace{-0.5em}
% \textcolor{red}{same comment as before: we do this example only with \PySIRTEM? I am okay with that but we should be upfron in saying why}

\begin{figure}[htbp]
  \centering

  % first row
  \begin{minipage}[b]{0.48\textwidth}
    \centering
    \includegraphics[width=\linewidth]{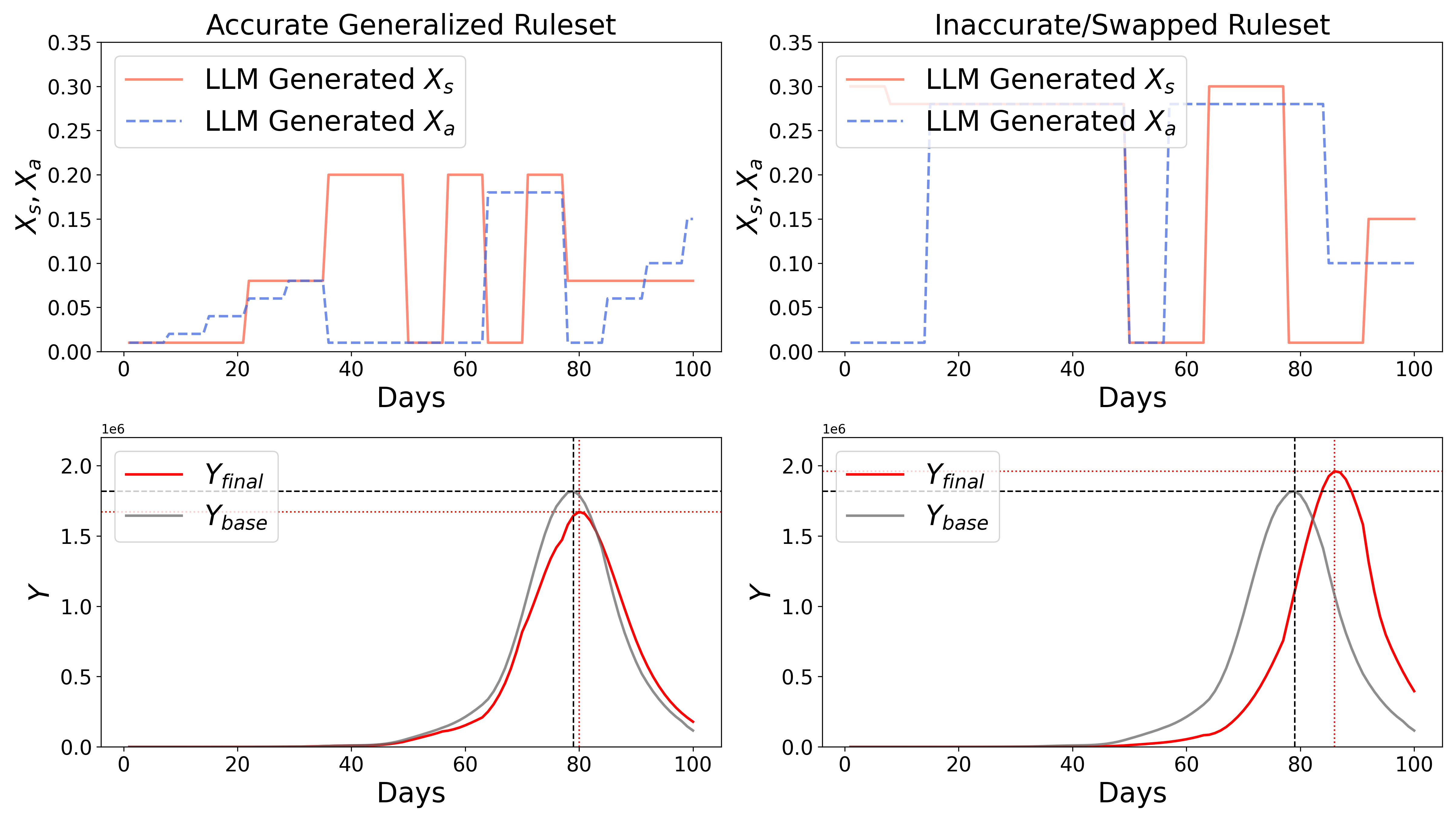}
  \end{minipage}\hfill
  \begin{minipage}[b]{0.48\textwidth}
    \centering
    \includegraphics[width=\linewidth]{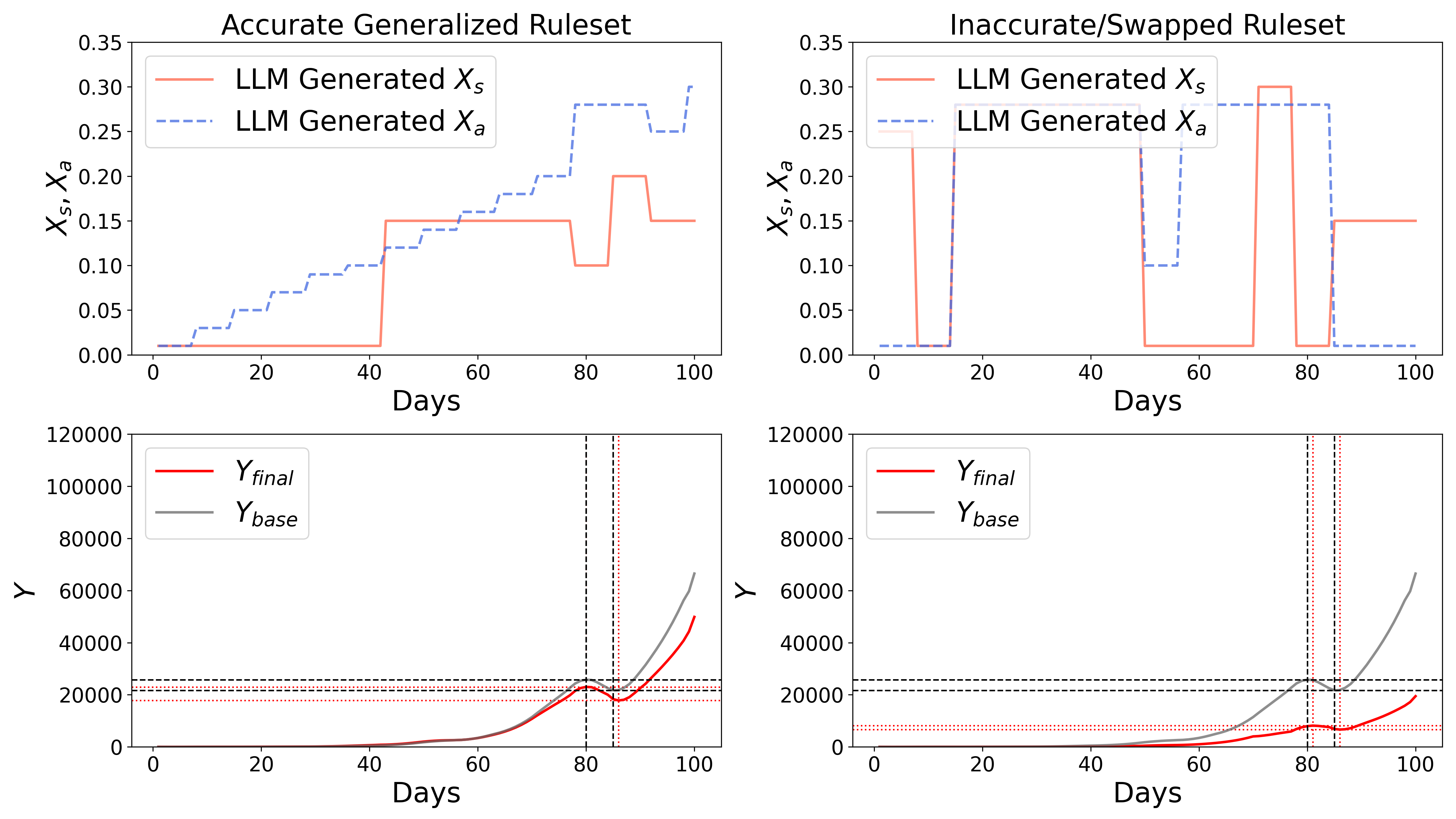}
  \end{minipage}
  \caption{Input Inference for an alternate (left) $Y^1_{\mbox{\tiny{base}}}$ with Prominent Phase Changes (right) $Y^2_{\mbox{\tiny{base}}}$ with Non-Prominent Phase Changes. E1 results are in the left subplot, E2 in the right subplot for each.}
  \label{fig:gen_base12}
\end{figure}

We assess whether the ruleset $\mathcal{R}_{\mbox{\tiny{final}}}$ inferred from a specific baseline $Y_{\mbox{\tiny{base}}}$ timeseries and cluster-derived contexts $\left((X_{s}^c, X_{a}^c), Y^c_{\mbox{\tiny{sim}}}\right)$
% \gp{also the context info I assume? can we describe exactly all the input that goes in here?}
, if utilized to construct input timeseries for a different $Y'_{\mbox{\tiny{base}}}$, produce a timeseries $Y'_{\mbox{\tiny{final}}}\approx Y'_{\mbox{\tiny{base}}}$. 
% For this study, we generated alternative baselines by setting desired infection and testing rates and executing the simulation on \PySIRTEM (creating a synthetic dataset). 
Due to space limitation, we perform this experiment only for our \PySIRTEM setup where we varied testing and infection rates to produce two $Y_{\mbox{\tiny{base}}}$ patterns with the following properties:
\begin{enumerate}[label=\roman*)]
    \setlength{\itemsep}{2pt}
    \setlength{\topsep}{1pt}
    \setlength{\parskip}{0pt}

    \item $Y^{1}_{\mbox{\tiny{base}}}$ with \textbf{prominent phase changes} that are easily mapped by the symbolic ruleset, but differing in timing (i.e., prolonged low initial phase, sharp peak no resurgence).

    \item $Y^{2}_{\mbox{\tiny{base}}}$ with \textbf{less distinct, non-prominent phase dynamics} that cannot be directly mapped to the ruleset (i.e. prolonged low initial phase, mild growth, slight dip, and late high resurgence).
\end{enumerate}
%\vspace{-0.5em}
For each baseline, we perform two experiments:
\begin{itemize}
    \setlength{\itemsep}{1pt}
    \setlength{\topsep}{1pt}
    \setlength{\parskip}{0pt}

    \item[(E1)] We provide the LLM with the ruleset but no clustering context (see Sec.~\ref{sec:context_sec}). The objective is to assess whether the ruleset inferred from another baseline \textit{alone} can generate inputs $\left( X_a,X_s\right)$ with acceptable accuracy for the new baselines $Y^{i}_{\mbox{\tiny{base}}}, i=1,2$.

    \item[(E2)] We modify the ruleset by inverting and replacing a subset of causal relationships, producing an \textbf{inaccurate} ruleset $\mathcal{R}_{\mbox{\tiny{incorrect}}}$, (provided in Appendix~\ref{app:supple}, Table~\ref{app:tab_inac}). We then examine how rule inaccuracies affect input generation and the alignment between $Y_{\mbox{\tiny{incorrect}}}$ and $Y_{\mbox{\tiny{base}}}$. 
\end{itemize}
%\vspace{-0.5em}
Fig.~\ref{fig:gen_base12} demonstrates the ability of the $\mathcal{R}_{\mbox{\tiny{final}}}$-induced inputs to infer trajectories that produce accurate outputs under both prominent and non-prominent phase dynamics (Fig.~\ref{fig:gen_base12}). In both cases, inputs generated under $\mathcal{R}_{\mbox{\tiny{final}}}$ capture the progression of $Y_{\mbox{\tiny{base}}}$ with reasonable accuracy. In contrast, under the modified ruleset $\mathcal{R}_{\mbox{\tiny{incorrect}}}$ we observe substantially greater misalignment—both temporally and in magnitude—between $Y_{\mbox{\tiny{incorrect}}}$ and $Y_{\mbox{\tiny{base}}}$. This supports the importance of a well-defined, structured ruleset for improved input generation.

% \begin{figure}[htbp]
%     \centering
%     \includegraphics[width=0.85\linewidth]{Figs/gen_base1.png}
%     \caption{Input Inference for an alternate $Y^1_{\mbox{\tiny{base}}}$ with Prominent Phase Changes. E1 results are in the left subplot, E2 results in the right subplot.}
%     \label{fig:gen_base1}
% \end{figure}

% \begin{figure}[t]
%     \centering
%     \includegraphics[width=0.85\linewidth]{Figs/gen_base2.png}
%     \caption{Input Inference for an alternate $Y^2_{\mbox{\tiny{base}}}$ with Non-Prominent Phase Changes. E1 results are in the left subplot, E2 results in the right subplot.}
%     \label{fig:gen_base2}
% \end{figure}

%\gp{
%\vspace{-0.5em}
\paragraph{Discussion and limitations.}While our evaluations demonstrate that the extracted rules are syntactically valid, semantically consistent, and effective in generating novel inputs given outputs, we acknowledge several limitations. First, our framework does not aim to recover the ``ground-truth'' generative rules of the simulator, instead constructing symbolic rules within a constrained logical grammar that provide valid and reusable causal explanations for observed trajectories. As such, traditional accuracy-based comparisons against latent mechanistic models are not applicable. Secondly, we do not benchmark against existing symbolic rule learners or interpretable predictors (e.g., shapelet-based classifiers or symbolic regression), as our objective is not prediction or classification, but symbolic abstraction of delayed causal relations in high-dimensional temporal simulations. 
% Future work may explore hybrid evaluation criteria that combine fidelity with parsimony and domain alignment. 
%}
%\vspace{-0.5em}
\section{Conclusions and Future Work}\label{sec::conclfw}
%\vspace{-0.5em}
We presented \epiExplain, a symbolic rule learning framework for causal abstraction in multivariate timeseries data. Unlike traditional forecasting or structural inference methods, \epiExplain constructs formal causal rules expressed in a constrained, human-interpretable symbolic language with explicit temporal and delay semantics. This structure enables LLMs to generate reusable explanations that link input trends to output phase behaviors without relying on domain-specific knowledge or direct simulator access. We demonstrate the approach across two distinct models and domains: an epidemic simulator mapping human-regulated testing rates to infection counts, and a building energy simulator mapping weather variables to electricity usage. Despite their differences in signal characteristics, periodicity, and causal structure, \epiExplain successfully extracts reusable symbolic rules that capture the governing temporal relationships. The closed-loop synthesis-and-verification process ensures rules are refined through simulation feedback, enabling robust alignment between symbolic explanations and system dynamics. We demonstrate the framework ability to generalize rules across diverse input trajectories and regenerate semantically consistent counterfactual inputs. The ruleset structured form makes it suitable for applications in counterfactual reasoning, treatment effect analysis, and simulation auditing.

% Future work will explore adaptive clustering strategies to improve the diversity and informativeness of reference trajectories, as well as fine-tuning LLMs to optimize rule refinement in the closed loop. We also aim to investigate agentic approaches that can autonomously propose and incorporate additional logical constraints during refinement, enabling more systematic and scalable rule inference. Furthermore, hybrid evaluation metrics that combine fidelity, simplicity, and semantic alignment will help benchmark symbolic causal frameworks like \epiExplain against emerging alternatives. 

Future work will explore adaptive clustering for more informative reference trajectories, fine-tuning LLMs for improved rule refinement, and agentic approaches to autonomously introduce logical constraints scalably and systematically. We also aim to develop hybrid evaluation metrics that balance fidelity, simplicity, and semantic alignment to benchmark symbolic causal frameworks like \epiExplain against emerging alternatives.

% \acks{Acknowledgements text.}

\bibliography{biblio}

\appendix

\section{Experimental Setup and Methodological Details}
% \update{
\subsection{\PySIRTEM Epidemic Simulator}
\label{app:pysirtem}
\PySIRTEM~\citep{biswas2025pysirtem} is an extended \textit{Susceptible-Expected-Infected-Recovered} (SEIR) model that uses Delay Differential Equations (DDEs) with compartments that represent, in addition to the original SEIR ones, health processes such as hospitalization, immunity, and testing. 
A complete list of transition parameters can be found in~\citep{sirtem}. Without loss of generality, this work uses \PySIRTEM as the basis to show the proposed explanation-generation approach. In fact, \epiExplain is independent from \PySIRTEM, but a simulator is needed in the evaluation loop (see Fig.~\ref{fig:model_overview}).

With \PySIRTEM as base simulator, to demonstrate \epiExplain, we focus on generating explanations for two timeseries, the \textit{symptomatic testing rate} ($X_s = \Phi_s= \left( x_{s,1}, x_{s,2}, x_{s,3}, \dots x_{s,T}\right)$) and the \textit{asymptomatic testing rate} ($X_a=\Phi_a= \left( x_{a,1}, x_{a,2}, x_{a,3}, \dots x_{a,T}\right)$), where $x_{s,t}$ and $x_{a,t}$ represent the daily ratio of the symptomatic and asymptomatic population tested against the total population for day $t$, respectively.
%} %All other relevant parameters are kept constant for this experiment. 
The pair $\left\lbrace \left(X_{s}, X_{a}\right)\right\rbrace_t$ can be modeled as a discrete time timeseries. 
These timeseries are important to understand the daily infected population (output timeseries) and they can be implemented as decisions (e.g., from the local government). Following the \PySIRTEM notation and the modeling introduced in~\citet{biswas2025pysirtem}, the following holds: 
%\vspace{-0.5em}
\begin{align}
    Y_t &= I_t = r \cdot \big( PS_t + PA_t + IA_t + ATN_t \big) + IS_t + STN_t.
    \label{eq:eq_inf}
\end{align}
In eqn.~\eqref{eq:eq_inf}, $PS$ and $PA$ are the pre-symptomatic and pre-asymptomatic population, $IS$ and $IA$ are the symptomatic and asymptomatic infected population, and $ATN$ and $STN$ are Asymptomatic and Symptomatic population who test negative (by error), respectively. The quantity $r$ is the ratio between transmission rates for the asymptomatic against symptomatic population. $PS, PA, IS, IA, STN,$ and $ATN$ are represented in \PySIRTEM as compartments and they are directly affected by the timeseries $X_s$ and $X_a$. 
%To set up the model for our experiments in this work, 
In this work, we treat \PySIRTEM as a black box system keeping all hyperparameters constant except the timeseries $X_s$ and $X_a$, which \epiExplain will produce as a prediction and explanation for the output timeseries $Y_t$. Namely, the relevant input-output pair for the causal inference analysis step of \epiExplain are $\left\lbrace(X_{s}, X_{a}), Y\right\rbrace$ where $X_{s}$ and $X_{s}$ are daily testing rates timeseries (symptomatic and asymptomatic, respectively) and $Y$ is the list of corresponding daily number of currently infected individuals. In this paper, we consider a simulation horizon of $T=100$ days, i.e., $t=1,2,\ldots,T$, and to maintain the assumptions in \PySIRTEM original work, we allow the timeseries for the test rates to only vary weekly, i.e., we impose the test rates to be stepwise constant functions. 

\epiExplain uses the \PySIRTEM module in two ways: (i) as a learning module (step 1 in Fig.~\ref{fig:model_overview}) to derive the baseline current daily infected cases $I_{base}$ from historical data, and secondly as a simulator (step 6 in Fig.~\ref{fig:model_overview}) to evaluate the effects of the timeseries inputs $(\Phi_s, \Phi_a)$. In this work, we refer to $\Phi_s, \Phi_a,$ and $I$ as $X_s, X_a, $and $Y$ respectively to avoid domain-specific notations.

\subsection{EnergyPlus Building Energy Simulator}
\label{energyPlus_sim}

EnergyPlus~\citep{crawley2001energyplus} is a physics-based building energy simulation tool that models building thermal dynamics, HVAC operation, internal loads, and environmental interactions using heat balance equations and system-level energy models. The simulator takes as input a building configuration file (IDF: \textit{Input Data File}) and a weather file (EPW: \textit{EnergyPlus Weather file}). 
The IDF file specifies the building geometry, dimensions, geological location, materials, HVAC systems, occupancy schedules, and other internal parameters. The IDF also contains the specifications for the run period, variables of interest, and frequency of reporting. The EPW file provides temporal weather data including dry-bulb temperature, solar radiation, humidity, wind speed, and other meteorological variables. EnergyPlus can operate at any temporal resolution. In our experiments, we use hourly simulation outputs.

EnergyPlus can model a wide range of outputs, including zone temperatures, heating and cooling loads, HVAC energy consumption, and total building electricity usage. Without loss of generality, we use EnergyPlus as a second simulation environment to demonstrate the domain-independence of \epiExplain. Similar to \PySIRTEM, \epiExplain treats EnergyPlus strictly as a black-box forward simulator within the evaluation loop (see Fig.~\ref{fig:model_overview}).

For the experiments in this work, we use the \textit{Supermarket} reference building IDF model provided within the standard EnergyPlus example files. As weather input, we use the file \texttt{USA\_VA\_Sterling-Washington\_Dulles.Intl.AP.epw} from the official EnergyPlus weather database. The simulation horizon is set to 100 consecutive hours and the baseline data is from Jan 1 to Jan 5, 1997. The relevant input–output pair for the causal analysis step of \epiExplain is 
\[
\left\lbrace (X_{d}, X_{g}), Y \right\rbrace,
\]
where:

\begin{itemize}[itemsep=2pt, topsep=2pt]
    \item $X_{d}$ denotes outdoor dry-bulb temperature (in $^\circ$C),
    \item $X_{g}$ denotes global horizontal irradiance (in Wh/m$^2$),
    \item $Y$ denotes total building electricity consumption (in J).
\end{itemize}

Outdoor dry-bulb temperature represents ambient air temperature affecting heat transfer across the building and Global horizontal irradiance measures the total solar radiation incident on a horizontal surface, influencing solar heat gains through windows and building surfaces. Both variables directly affect cooling loads and HVAC operation and impact electricity consumption which is our output variable of interest.

In this work, we observe temperature and irradiance as multivariate input trajectories $X_t$, while keeping all building physical properties and HVAC parameters fixed. The simulator maps these environmental inputs to hourly electricity consumption $Y$. 

\epiExplain uses the EnergyPlus module analogously to the epidemic case: (i) to generate a baseline electricity consumption trajectory $Y_{\text{base}}$ under reference weather conditions, and (ii) to evaluate candidate multivariate input trajectories during rule refinement. As before, we abstract the domain-specific notation and refer to temperature, irradiance, and electricity consumption as $X_{d}$, $X_{g}$, and $Y$, respectively, to maintain consistency across domains.

\subsection{Dimensionality Reduction and Clustering}\label{app::dredClust}
To reduce redundancy and extract representative data points that will form the training set for the explanation generation step (Sec.~\ref{sec::expInf}), we apply dimensionality reduction and clustering on the set of timeseries resulting from the input timeseries learning step (Sec.~\ref{sec::TRlearn}), $\{\left(X_s^p, X_a^p\right)\}^{N}_{p=1}$. 
%\vspace{-0.5em}
\paragraph{Dimensionality reduction.}As each candidate comprises of two daily timeseries vectors, $X_s = \left(x_{s, 1}, x_{s, 2}, \dots x_{s, T}\right)$ and $X_a = \left(x_{a, 1}, x_{a,2}, \dots, x_{a, T}\right)$, the input space is $2T$-dimensional. Because of the high-dimensionality and non-linearity of the input space, we first reduce the dimensionality considering only the weekly changing points (this is not necessary to the application of the framework), and, on the reduced dimensionality timeseries, we employ the Uniform Manifold Approximation and Projection (UMAP)~\citep{umap2020} technique to represent the data points in a 2-D space embedding for clustering. %More specifically, we employ 
UMAP %\update{
%as 
%it 
facilitates visual confirmation of cluster separability %}. UMAP constructs 
by constructing a high-dimensional topological representation of the data and then optimizing a low-dimensional graph to be as structurally similar as possible through cross-entropy minimization. Unlike other linear (Principle Component Analysis or PCA) and non-linear techniques (t-distributed Stochastic Neighbor Embedding or t-SNE), UMAP preserves both local and global structure with competitive accuracy~\citep{umap2020}. 

%\vspace{-0.5em}
\paragraph{Clustering the reduced-dimensionality timeseries.}Upon reduction to a 2-D input space, we employ a $K$-Means clustering to identify distinct input timeseries ``trends''. %\update{
We choose $K$-Means clustering due to its simplicity and efficiency in detecting separable clusters. The input timeseries corresponding with the centroids of each cluster are represented by $\left(X_{s}^c, X_{a}^c\right)$. As all $\left(X_{s}^c, X_{a}^c\right)$ result in output trends identical to $Y_{base}$, we denote the input-output pairs of each clusters as $\left((X_{s}^c, X_{a}^c),Y_{base}\right)$ for simplicity.%}

The primary purpose of clustering in \epiExplain is to elect a single representative data point from each solution cluster as inducing timeseries within the training set during the LLM phase (see Sec.~\ref{sec:LLM}). We argue that such data reduction can help in the training context emphasizing diversity, removing redundant examples that may result into overly ``engineered'' explanations (rulesets $\mathcal{R}$), while removing the influence of individual cluster size or density. Since all the input pairs generate the same output trajectory ($Y_{\mbox{\tiny{base}}}$), we argue that a diverse training set facilitates the LLM task to extrapolate the implicit, causal relation between the variables. 

\begin{figure}[htbp]
    \centering
    \includegraphics[width=0.7\linewidth]{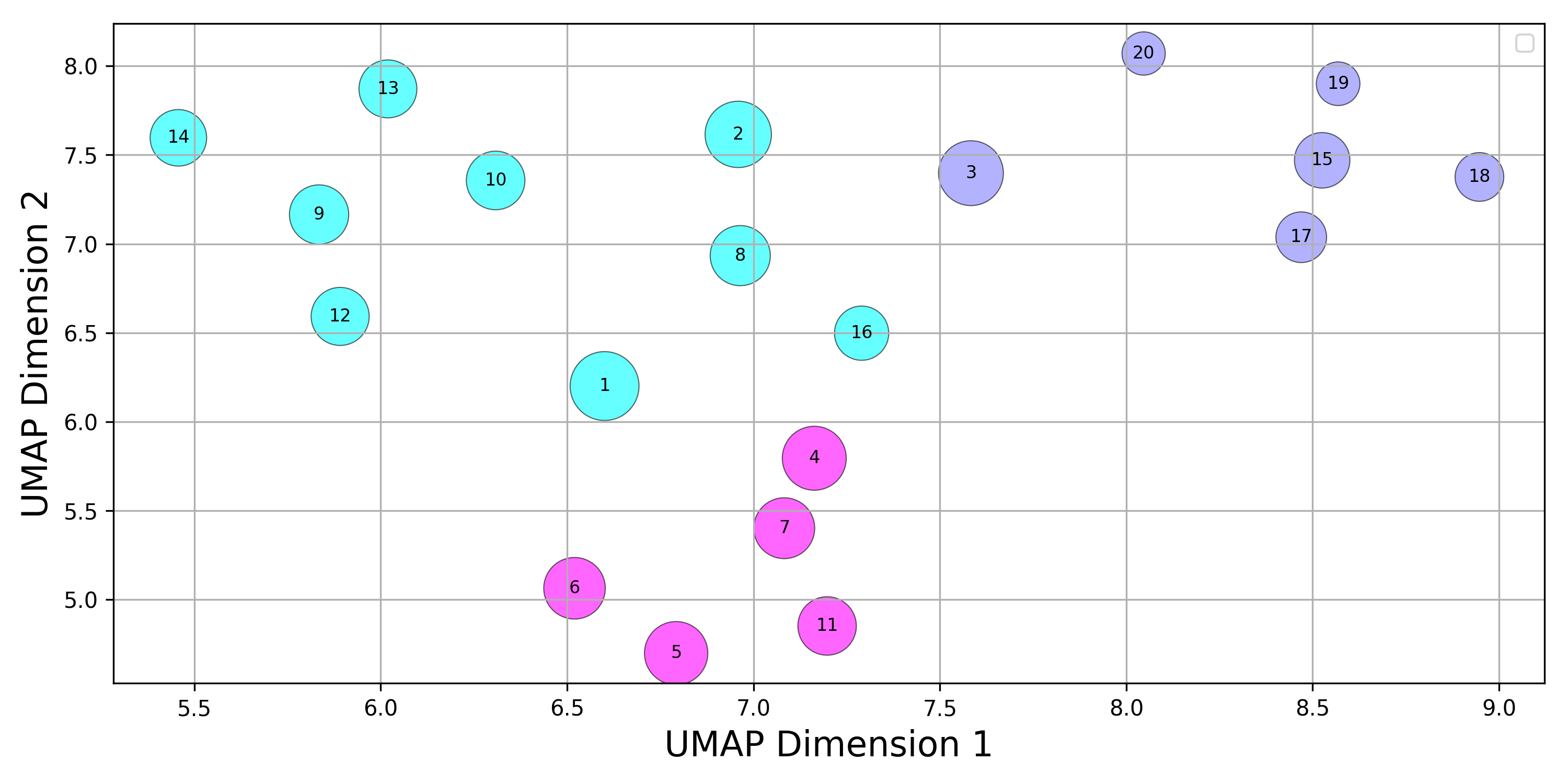}
    \caption{Clustering of the initial solution pool $\left\lbrace (X^p_s, X^p_a)\right\rbrace_p$ from solving $P_{\mbox{\tiny{LEARN}}}(p)$ for \PySIRTEM setup.}
    \label{fig:clusters}
\end{figure}

\section{Extended Results and Analysis}
\label{app:causal_ruleset_sec}
In this section, we report the final extracted ruleset after the closed-loop refinement process of \epiExplain for both \PySIRTEM (Table.~\ref{tab:causal_rules}) and EnergyPlus (Table~\ref{tab:causal_rules_eng}).
\begin{table*}[htbp]
\centering
\scriptsize
\renewcommand{\arraystretch}{1.1} % Slightly reduced row height
\caption{Causal rules extracted from timeseries examples and context for \PySIRTEM Setup.}
\setlength{\tabcolsep}{3pt}       % Reduce horizontal padding in cells
\begin{tabular}{p{0.04\linewidth}|p{0.07\linewidth}|p{0.11\linewidth}|p{0.10\linewidth}|p{0.07\linewidth}|p{0.13\linewidth}|p{0.33\linewidth}}
\toprule
\textbf{Rule ID} & \textbf{Input Window} & \textbf{$X_s$ Trend} & \textbf{$X_a$ Trend} & \textbf{Output Window} & \textbf{$Y$ Phase} & \textbf{Symbolic Rule} \\
\hline
R1 & [0, 14] & Low & SlowRise & [15, 29] & InitialGrowth & 
$\Box_{[0,14]}(\mathrm{Low}(X_s) \wedge \mathrm{SlowRise}(X_a)) \rightarrow \Diamond_{[15,29]}\mathrm{InitialGrowth}(Y)$ \\
\hline
R2 & [14, 28] & Moderate & SlowRise & [29, 43] & PeakFormation & 
$\Box_{[14,28]}(\mathrm{Moderate}(X_s) \wedge \mathrm{SlowRise}(X_a)) \rightarrow \Diamond_{[29,43]}\mathrm{PeakFormation}(Y)$ \\
\hline
R3 & [28, 42] & SharpSpike & Low & [43, 57] & PostPeak & 
$\Box_{[28,42]}(\mathrm{SharpSpike}(X_s) \wedge \mathrm{Low}(X_a)) \rightarrow \Diamond_{[43,57]}\mathrm{PostPeak}(Y)$ \\
\hline
R4 & [42, 56] & DipThenRise & Low & [57, 71] & Decline & 
$\Box_{[42,56]}(\mathrm{DipThenRise}(X_s) \wedge \mathrm{Low}(X_a)) \rightarrow \Diamond_{[57,71]}\mathrm{Decline}(Y)$ \\
\hline
R5 & [56, 70] & SlowRise & HighStable & [71, 85] & Resurgence & 
$\Box_{[56,70]}(\mathrm{SlowRise}(X_s) \wedge \mathrm{HighStable}(X_a)) \rightarrow \Diamond_{[71,85]}\mathrm{Resurgence}(Y)$ \\
\hline
R6 & [70, 100] & Moderate & SlowRise & [85, 100] & PeakFormation & 
$\Box_{[70,100]}(\mathrm{Moderate}(X_s) \wedge \mathrm{SlowRise}(X_a)) \rightarrow \Diamond_{[85,100]}\mathrm{PeakFormation}(Y)$ \\
\bottomrule
\end{tabular}
\label{tab:causal_rules}
\end{table*}

\begin{table*}[htbp]
\centering
\scriptsize
\renewcommand{\arraystretch}{1.1} % Slightly reduced row height
\caption{Causal rules extracted from timeseries examples and context for EnergyPlus Setup.}
\setlength{\tabcolsep}{3pt}       % Reduce horizontal padding in cells
\begin{tabular}{p{0.04\linewidth}|p{0.07\linewidth}|p{0.11\linewidth}|p{0.10\linewidth}|p{0.07\linewidth}|p{0.13\linewidth}|p{0.33\linewidth}}
\toprule
\textbf{Rule ID} & \textbf{Input Window} & \textbf{$X_{t,1}$ Trend} & \textbf{$X_{t,2}$ Trend} & \textbf{Output Window} & \textbf{$Y$ Phase} & \textbf{Symbolic Rule} \\
\hline
R1 & [0, 9] & Low & Low & [0, 8] & Decline & 
$\Box_{[0,9]}(\mathrm{Low}(X_{t,1}) \wedge \mathrm{Low}(X_{t,2})) \rightarrow \Diamond_{[0,8]}\mathrm{Decline}(Y)$ \\
\hline
R2 & [9, 18] & SlowRise & SharpSpike & [9, 17] & InitialGrowth & 
$\Box_{[9,18]}(\mathrm{SlowRise}(X_{t,1}) \wedge \mathrm{SharpSpike}(X_{t,2})) \rightarrow \Diamond_{[9,17]}\mathrm{InitialGrowth}(Y)$ \\
\hline
R3 & [18, 21] & DipThenRise & LowModerate & [18, 20] & PostPeak & 
$\Box_{[18,21]}(\mathrm{DipThenRise}(X_{t,1}) \wedge \mathrm{LowModerate}(X_{t,2})) \rightarrow \Diamond_{[18,20]}\mathrm{PostPeak}(Y)$ \\
\hline
R4 & [21, 30] & Low & Low & [21, 29] & Decline & 
$\Box_{[21,30]}(\mathrm{Low}(X_{t,1}) \wedge \mathrm{Low}(X_{t,2})) \rightarrow \Diamond_{[21,29]}\mathrm{Decline}(Y)$ \\
\hline
R5 & [30, 42] & Moderate & HighStable & [30, 41] & PeakFormation & 
$\Box_{[30,42]}(\mathrm{Moderate}(X_{t,1}) \wedge \mathrm{HighStable}(X_{t,2})) \rightarrow \Diamond_{[30,41]}\mathrm{PeakFormation}(Y)$ \\
\hline
R6 & [42, 48] & DipThenRise & LowModerate & [42, 47] & PostPeak & 
$\Box_{[42,48]}(\mathrm{DipThenRise}(X_{t,1}) \wedge \mathrm{LowModerate}(X_{t,2})) \rightarrow \Diamond_{[42,47]}\mathrm{PostPeak}(Y)$ \\
\hline
R7 & [48, 54] & Low & Low & [48, 53] & Decline & 
$\Box_{[48,54]}(\mathrm{Low}(X_{t,1}) \wedge \mathrm{Low}(X_{t,2})) \rightarrow \Diamond_{[48,53]}\mathrm{Decline}(Y)$ \\
\hline
R8 & [54, 66] & Moderate & HighStable & [54, 65] & PeakFormation & 
$\Box_{[54,66]}(\mathrm{Moderate}(X_{t,1}) \wedge \mathrm{HighStable}(X_{t,2})) \rightarrow \Diamond_{[54,65]}\mathrm{PeakFormation}(Y)$ \\
\hline
R9 & [66, 72] & DipThenRise & LowModerate & [66, 71] & PostPeak & 
$\Box_{[66,72]}(\mathrm{DipThenRise}(X_{t,1}) \wedge \mathrm{LowModerate}(X_{t,2})) \rightarrow \Diamond_{[66,71]}\mathrm{PostPeak}(Y)$ \\
\hline
R10 & [72, 78] & Low & Low & [72, 77] & Decline & 
$\Box_{[72,78]}(\mathrm{Low}(X_{t,1}) \wedge \mathrm{Low}(X_{t,2})) \rightarrow \Diamond_{[72,77]}\mathrm{Decline}(Y)$ \\
\hline
R11 & [78, 90] & SlowRise & HighStable & [78, 89] & Resurgence & 
$\Box_{[78,90]}(\mathrm{SlowRise}(X_{t,1}) \wedge \mathrm{HighStable}(X_{t,2})) \rightarrow \Diamond_{[78,89]}\mathrm{Resurgence}(Y)$ \\
\hline
R12 & [90, 96] & DipThenRise & LowModerate & [90, 95] & PostPeak & 
$\Box_{[90,96]}(\mathrm{DipThenRise}(X_{t,1}) \wedge \mathrm{LowModerate}(X_{t,2})) \rightarrow \Diamond_{[90,95]}\mathrm{PostPeak}(Y)$ \\
\hline
R13 & [96, 102] & Low & Low & [96, 100] & Decline & 
$\Box_{[96,102]}(\mathrm{Low}(X_{t,1}) \wedge \mathrm{Low}(X_{t,2})) \rightarrow \Diamond_{[96,100]}\mathrm{Decline}(Y)$ \\
\bottomrule
\end{tabular}
\label{tab:causal_rules_eng}
\end{table*}

\begin{figure}[htbp]
  \centering

  % first row
  \begin{minipage}[b]{0.48\textwidth}
    \centering
    \includegraphics[width=\linewidth]{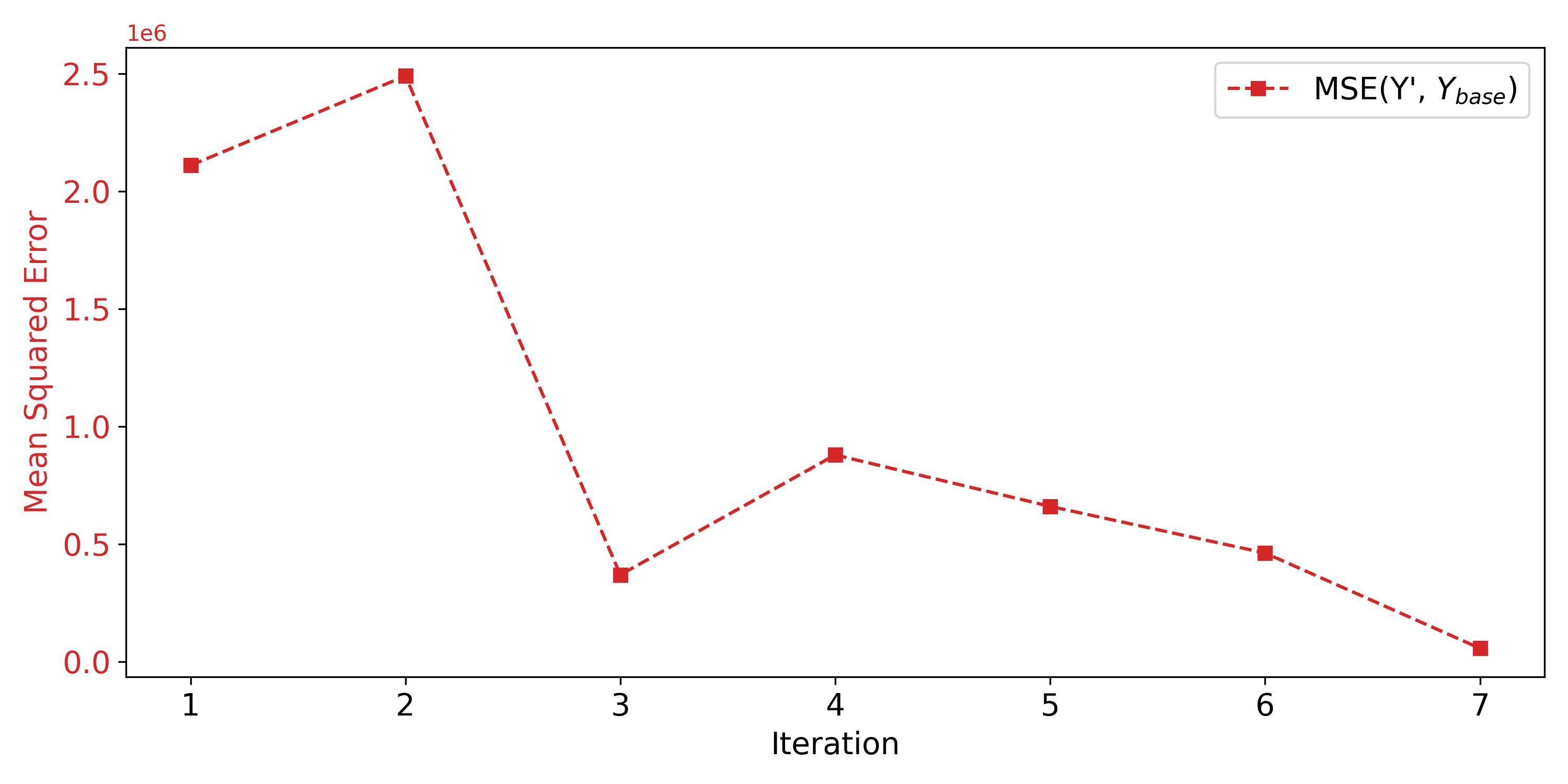}
  \end{minipage}
  \begin{minipage}[b]{0.48\textwidth}
  \centering
    \includegraphics[width=\linewidth]{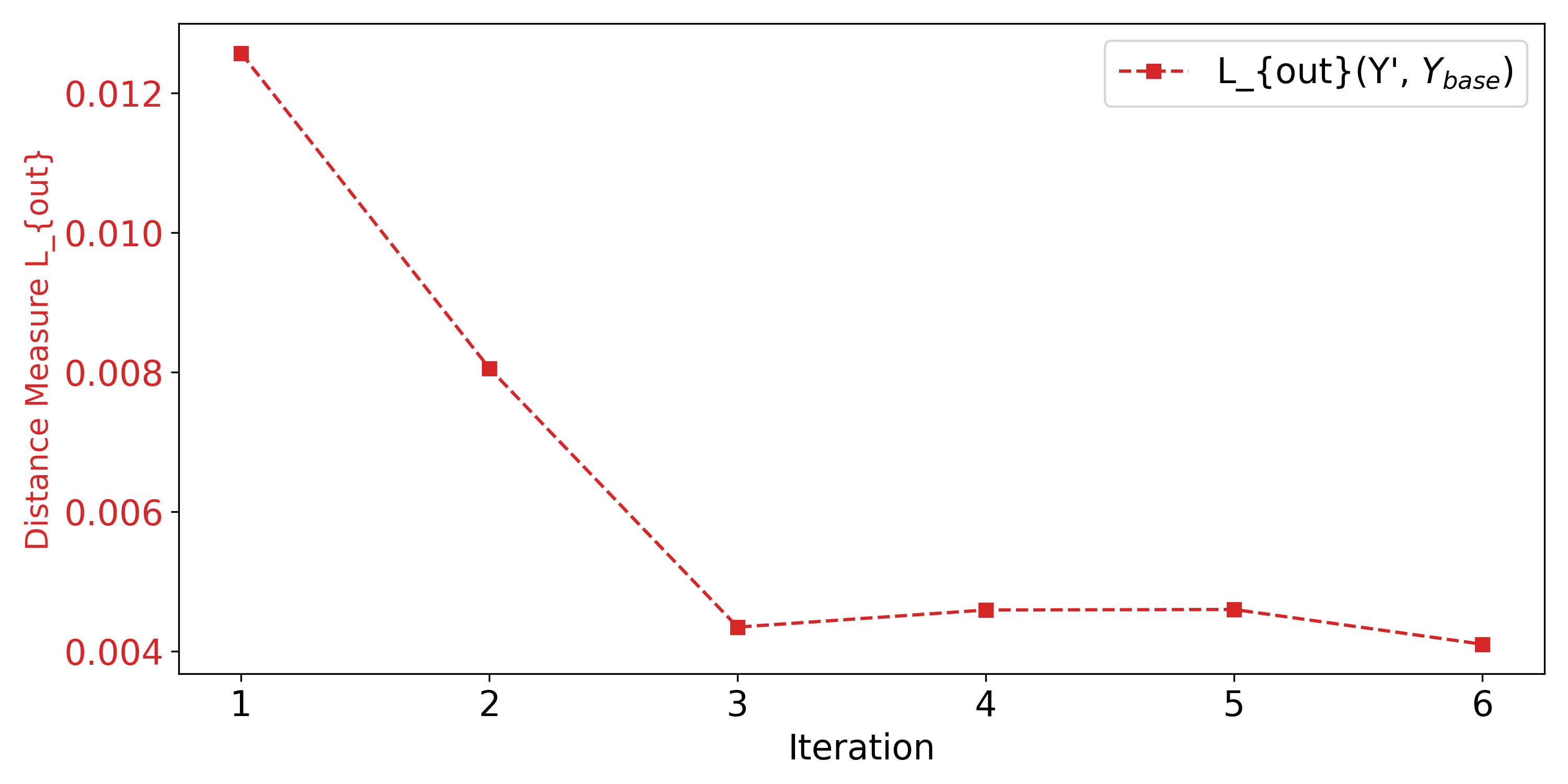}
  \end{minipage}
  \caption{\epiExplain input timeseries produced Mean Squared Difference across iterations for (left) \PySIRTEM (right) EnergyPlus}
  \label{fig:RQ1}
\end{figure}

Figure~\ref{fig:RQ1} reports the iterative reconstruction loss produced by the closed-loop pipeline. At each iteration, a symbolic ruleset is extracted or refined, the LLM is prompted to generate input trajectories consistent with the rules, and the simulator is executed to obtain $Y_{\text{sim}}$. The output loss $L_{\text{out}}(Y_{\text{sim}}, Y_{\text{base}})$—measured as Mean Squared and Normalized Mean Squared Difference—is then computed. The decreasing trend observed for both (a) \PySIRTEM and (b) EnergyPlus demonstrates that successive rule refinements generate inputs whose simulated outputs progressively align with the baseline trajectory, highlighting the effectiveness of the iterative refinement--verification process.

\section{Detailed Prompts for LLM}
\label{app:supple}
The codebase used during the course of this work can be found \href{https://github.com/Preetom1905011/CausalSeriesExplain}{here}.
\subsection{\epiExplain Framework Prompts}
The following prompts are used during \epiExplain's LLM-based Explanation Generation (Sec.~\ref{sec:LLM}): 
\begin{enumerate}
    \item \textbf{Initial Explanation Inference Prompt}. This prompt provides the LLM with cluster-derived contexts and asks to generate an interpretable, formal ruleset describing input-output relations (Fig.~\ref{fig:prompt_init}).
    \item \textbf{Input set generation using ruleset.} This prompt instructs the LLM to use the provided ruleset and cluster-derived context to generate a novel input series (Fig.~\ref{fig:prompt_input_gen}).
    \item \textbf{Ruleset refinement prompt.} This provides the LLM with the simulated output $Y'$ resulting from the generated $(X_s', X_a')$ and $Y_{base}$ along with the previously available contexts. The LLM is then asked to assess the deviations of $Y'$ and $Y_{base}$ and refine the ruleset (Fig.~\ref{fig:prompt_refine_rule}).
\end{enumerate}

\begin{figure}[htbp]
\centering
\scriptsize
\begin{tcolorbox}[title=Goal: Generate initial causal rules \(R_0\) from clustering context]
\textbf{Prompt:} \\

You are given several input-output examples from a timeseries simulation model. Each example consists of a multivariate input timeseries $X_t$ and a univariate daily output timeseries $Y$.

The input $X_t$ is defined over a time interval length $\{\Delta\}$. If $Y$ spans 100 days, then $X_t$ will contain $\lceil 100 / \Delta \rceil$ interval values. The relationship between inputs and outputs is be \{direct/inverse\} and may include temporal lag/delay effects.

Your goal is to observe the input-output relationships from the examples and extract symbolic causal rules of the form:

``If trends in selected components of $X_t$ persist over a given time interval, then the output $Y$ will exhibit a specific phase after a delay.''

Please divide the baseline output into specific output phases listed below in order of timing and infer the multivariate input trends that are responsible. Each output phase should be unique (no duplicate entries in the output column), and the rules should provide input and corresponding output trend behavior for the total time window (i.e., all input windows should concatenate to cover from $T=0$ to $T=100$). Input time windows must be multiples of $\Delta$.

Fill in the table below with symbolic causal rules. Each rule should specify:

\begin{itemize}[itemsep=0.25pt, topsep=0.25pt]
    \item Input Trend Window: the time interval where a specific trend in selected components of $X_t$ is observed.
    \item Trend Description: qualitative trends over that window.
    \item Output Phase Window: the time interval where a specific phase in $Y$ occurs, as a delayed effect of the input.
    \item Phase Description: the qualitative behavior of $Y$.
    \item Symbolic Rule: a formal statement capturing this as a logical implication.
\end{itemize}

Use the vocabulary:

\begin{itemize}[itemsep=0.2pt, topsep=0.2pt]
    \item Trends: Low, Moderate, HighStable, SharpSpike, SlowRise, DipThenRise, LowModerate
    \item Output Phases: InitialGrowth, PeakFormation, PostPeak, Decline, Resurgence
\end{itemize}

Below is the list of example input-output points and Baseline Expected Output:\\
---- Example 0 ------\\
\quad $X_{t,1}$: [0.018, 0.14, ... $K$ values], 
\quad $X_{t,2}$: [0.06, 0.23, ... $K$ values], 
$Y$:  [15, 12, ... $T$ values]\\
---- Example 1 ------\\
\quad $X_{t,1}$: [0.025, 0.083, ... $K$ values], 
\quad $X_{t,2}$: [0.05, 0.19, ... $K$ values], 
$Y$:  [15, 12, ... $T$ values]\\
\textit{\{Insert More ...\}}\\
—---------------------\\
Baseline $Y$:  [15, 12, ... $T$ values]\\
Table format:
\begin{center}
\scriptsize
\begin{tabular}
{|p{0.04\linewidth}|p{0.08\linewidth}|p{0.09\linewidth}|p{0.08\linewidth}|p{0.08\linewidth}|p{0.13\linewidth}|p{0.30\linewidth}|}
\hline
\textbf{Rule ID} & \textbf{Input Trend Window} & \textbf{$X_s$ Trend} & \textbf{$X_a$ Trend} & \textbf{Output Phase Window} & \textbf{$Y$ Phase} & \textbf{Symbolic Rule} \\
\hline
R1 & [0, 15] & Low & SlowRise & [0, 20] & PeakFormation & $\Box_{[0,15]}(\text{Low}(X_s) \land \text{SlowRise}(X_a)) \rightarrow \Diamond_{[0,20]}\text{PeakFormation}(Y)$ \\
\hline
R2 & [15, 30] & Low & Low & [20, 35] & InitialGrowth & $\Box_{[15, 30]}(\text{Low}(X_s) \land \text{Low}(X_a)) \rightarrow \Diamond_{[20, 35]}\text{InitialGrowth}(Y)$ \\
\hline
\end{tabular}
\end{center}

Make sure to use the table format shown above. Do not use domain-specific terminology; instead, describe timeseries patterns using abstract trend labels. Ensure the symbolic rule column uses the logical form:

\[
\Box_{[t1,t2]}(\text{Trend}(X_{t,i}) \land \text{Trend}(X_{t,j}) \land \dots) \rightarrow \Diamond_{[t3,t4]}\text{Phase}(Y)
\]

Ensure:
\begin{itemize}[itemsep=0.25pt, topsep=0.25pt]
    \item All time windows are valid.
    \item Input windows are multiples of $\Delta$.
    \item Output phases are unique and cover the full timeline.
    \item The symbolic rule strictly follows the required logical template.
    \item Only abstract trend labels are used.
\end{itemize}

\end{tcolorbox}
\caption{Prompt for generating the initial ruleset.}
\label{fig:prompt_init}
\end{figure}

\begin{figure}[h]
\centering
\scriptsize
\begin{tcolorbox}[title={Goal: Novel input set $(X_{s}', X_{a}')$ generation using ruleset $\mathcal{R}_i$}]

\textbf{Prompt:} You are given several input-output examples from a timeseries simulation model and a causal ruleset describing input trends and their corresponding output effects.

Each example consists of:
\begin{itemize}[itemsep=0pt, topsep=0pt]
    \item A multivariate input timeseries $X_t$ defined over interval length $\Delta$
    \item A daily output timeseries $Y$
\end{itemize}

All example inputs generate the same output $Y$. The relationship between inputs and outputs is direct and may include temporal lag/delay effects. The baseline output $Y$ spans 100 days. The input $X_t$ is defined over interval length $\{\Delta\}$, so $X_t$ contains $\lceil 100 / \Delta \rceil$ interval values.

Input constraints:
\begin{itemize}[itemsep=0.25pt, topsep=0.25pt]
    \item All $X_{t,1}$ values must lie in the range \{\textit{Insert Range}\}
    \item All $X_{t,2}$ values must lie in the range \{\textit{Insert Range}\}
\end{itemize}

You must use the provided ruleset to generate a novel multivariate input sequence $X_t$ that is different from all example inputs but produces an output $Y$ that closely follows the baseline $Y$ trend.

\textbf{Instructions:}

\begin{enumerate}[itemsep=0.25pt, topsep=0.25pt]
    \item Analyze the baseline $Y$ trend and the provided ruleset.
    \item Divide the baseline $Y$ into output phases.
    \item For each output phase:
    \begin{itemize}[itemsep=0.2pt, topsep=0.2pt]
        \item Identify the corresponding rule in the ruleset.
        \item Determine the required input trends.
    \end{itemize}
    \item Generate a new multivariate input sequence $X_t$:
    \begin{itemize}[itemsep=0.2pt, topsep=0.2pt]
        \item Length must match the number of intervals determined by $\Delta$.
        \item Trends must be consistent with the ruleset.
        \item The full timeline must be covered.
        \item The trend pattern must differ from all previous example inputs.
        \item All input values must satisfy the specified bounds.
    \end{itemize}
    \item For each output phase, briefly justify your input choices for the corresponding input intervals, referencing:
    \begin{itemize}[itemsep=0.2pt, topsep=0.2pt]
        \item The applied rule
        \item The delayed and direct relationship between inputs and output
    \end{itemize}
\end{enumerate}

Below is the list of example input-output points and Baseline Expected Output:\\
\textit{\{Insert Clustering Context ...\}}\\
—---------------------\\
Baseline $Y$:  \{Baseline Output\}\\
Rules Table: $<\textit{Insert Ruleset in Markdown format}>$\\
\textbf{Required Output Format:}

\begin{enumerate}[itemsep=0.2pt, topsep=0.2pt]
    \item \textbf{Generated Input Series:} Provide the full multivariate input sequence:\\
    - $X_{t,1}$: $[v_1, v_2, \dots, v_K]$, $X_{t,2}$: $[v_1, v_2, \dots, v_K]$, $\dots$

    \item \textbf{Phase-wise Justification:} For each output phase, provide: \\
    \textbf{Output Phase:} PhaseName,
    \textbf{Output Window:} $[t_3, t_4]$,
    \textbf{Applied Rule:} RuleID,
    \textbf{Required Input Trend Window:} $[t_1, t_2]$,
    \textbf{Assigned Input Trends:} $X_{t,i}$: Trend \\
    \textbf{Justification:} Brief explanation referencing the rule and the delayed/direct or inverse relationship.
    
\end{enumerate}

Ensure:
\begin{itemize}[itemsep=0.2pt, topsep=0.2pt]
    \item The generated input trends differ structurally from all example inputs.
    \item Input intervals are multiples of $\Delta$.
    \item The ruleset is followed consistently.
    \item The entire timeline is covered.
    \item All input values lie within the specified range.
    \item The resulting $Y$ closely follows the baseline $Y$.
\end{itemize}

\end{tcolorbox}
\caption{Prompt for Novel Input series generation using inferred ruleset $\mathcal{R}_i$.}
\label{fig:prompt_input_gen}
\end{figure}

\begin{figure}[htbp]
\centering
\scriptsize
\begin{tcolorbox}[title={Goal: Ruleset Refinement by evaluating \(Y'\) (generated from \((X_{s}', X_a')\) and \(Y_{base}\)}]
\textbf{Prompt:} \\

You previously generated a multivariate input series $X_t'$ using a causal ruleset, simulated the system, and obtained an output $Y'$. The simulated output $Y'$ does not closely match the Baseline $Y$.

The relationship between inputs and output is \{direct/inverse\} and delayed. This is critical and must guide your diagnosis and rule updates.

\textbf{Generated Input:} \\

$X_t'$:
Your generated weekly input timeseries:\\
\{\textit{Insert $X_{t,sim}$}\}

corresponding $Y'$: 
\{\textit{Insert $Y_{sim}$}\}\\
—---------------------\\
Baseline $Y$: \{\textit{Insert $Y_{base}$}\}\\
Previous Rules Table: $<\textit{Insert Ruleset in Markdown Format}>$\\
---\\\textbf{Instructions:} \\

1. Divide the Baseline $Y$ into distinct output phases in chronological order. \\
2. Compare $Y'$ with Baseline $Y$ and identify mismatches, including:
Timing shifts, magnitude differences, shape differences (peaks, valleys, slopes), and phase misalignment. \\
3. Diagnose which input trends in $X_t'$ caused the mismatch, considering whether they were too high or too low, too early or too late, too sharp or too flat, not sustained long enough, or incorrectly capturing the direct or delayed relationship. \\
4. Update the ruleset to better guide future input generation. \\

\textbf{Constraints for Updated Rules:} \\

Each output phase must be unique. \\
Input windows must be multiples of $\Delta$. \\
Input windows must concatenate to cover the full timeline ($T = 0$ to $T = 100$). \\
Use only abstract trend labels. \\
Do not use domain-specific terminology. \\
When appropriate, refine only the input variable most responsible for a mismatch. \\

\textbf{Allowed Vocabulary:} \\

Input Trends: Low, Moderate, HighStable, SharpSpike, SlowRise, DipThenRise, LowModerate \\
Output Phases: InitialGrowth, PeakFormation, PostPeak, Decline, Resurgence \\

\textbf{Required Output Format:} \\
Return the updated rules in following Markdown table format:

\begin{center}
\scriptsize
\begin{tabular}
{|p{0.04\linewidth}|p{0.08\linewidth}|p{0.09\linewidth}|p{0.08\linewidth}|p{0.08\linewidth}|p{0.13\linewidth}|p{0.30\linewidth}|}
\hline
\textbf{Rule ID} & \textbf{Input Trend Window} & \textbf{$X_s$ Trend} & \textbf{$X_a$ Trend} & \textbf{Output Phase Window} & \textbf{$Y$ Phase} & \textbf{Symbolic Rule} \\
\hline
R1 & [t1,t2] & Low & SlowRise & [t3,t4] & PeakFormation & $\Box_{[t1,t2]}(\text{Low}(X_{t,1}) \land \text{SlowRise}(X_{t,2})) \rightarrow \Diamond_{[t3,t4]}\text{PeakFormation}(Y)$ \\
\hline
\end{tabular}
\end{center}

\textbf{Symbolic Rule Format:} \\

Each rule must follow:

\[
\Box_{[t1,t2]}(\text{Trend}(X_{t,i}) \land \text{Trend}(X_{t,j}) \land \dots) \rightarrow \Diamond_{[t3,t4]}\text{Phase}(Y)
\]

Ensure:
Rules better explain the mismatch between $Y'$ and Baseline $Y$. The direct and delayed relationship must be properly encoded. The updated rules should improve future alignment with Baseline $Y$.

\end{tcolorbox}

\caption{Prompt for Ruleset Refinement.}
\label{fig:prompt_refine_rule}
\end{figure}

\subsection{Validation Experiment Prompts}
\label{app:val_prompts}
The following prompts have been used to validate \epiExplain:
\begin{enumerate}
    \item \textbf{(S1) Executing LLM with no rules and no cluster-derived context (Sec.~\ref{sec:context_sec}).} we provide the LLM with only the target output $Y_{\mbox{\tiny{base}}}$, existence of causal dependency between inputs and outputs, and the valid input ranges. The LLM is further instructed to avoid any domain specific knowledge (Fig~\ref{fig:prompt_no_rule_no_cont}).
    \item \textbf{(E2) Input generation with inaccurate ruleset.} We modify the ruleset by inverting and replacing a subset of the input-output timeseries relationships. This produces an \textbf{inaccurate} ruleset $\mathcal{R}_{\mbox{\tiny{incorrect}}}$ (Table~\ref{app:tab_inac}), and we want to observe how the inaccuracy of rules affects the generation of the input timeseries and, consequently, the alignment between $Y_{\mbox{\tiny{incorrect}}}$ and $Y_{\mbox{\tiny{base}}}$. We use the inaccurate ruleset in the same prompt as Fig.~\ref{fig:prompt_input_gen}.
    
\end{enumerate}

\begin{figure}[htbp]
\centering
\begin{tcolorbox}[title={Goal: Novel Input Generation without Rules or Context}]
\footnotesize
\textbf{Prompt:} \\

You are given a system that takes as input a multivariate timeseries parameters $X_{t}$ defined over interval length $\Delta$, and produces a daily output timeseries $Y$. If $Y$ spans \{T\} days, each input sequence contains $\lceil \{T\} / \{\Delta\} \rceil$ values.

Given a target output trajectory $Y$ over \{T\} days and the valid range for each input parameter, your task is to infer a pair of input sequences that could plausibly generate the same output.

Input constraints:
\begin{itemize}[itemsep=0.2pt, topsep=0.2pt]
    \item All $X_{t,1}$ values must lie within $\{ \textit{Insert Range}_1 \}$
    \item All $X_{t,2}$ values must lie within $\{ \textit{Insert Range}_2 \}$ \\
    $\{\textit{Insert More}\}$
\end{itemize}

Do not provide code. Infer the input values directly from the given output pattern. Do not use any domain-specific knowledge; treat the inputs and outputs as abstract functional relationships.

\textbf{Required Output Format:} \\
Provide the full multivariate input sequences explicitly:
\[
X_{t,1} = [v_1, v_2, \dots, v_K], \quad
X_{t,2} = [v_1, v_2, \dots, v_K].
\]
\end{tcolorbox}
\caption{Prompt for Input Generation with \textbf{no rules + no clustering context}}
\label{fig:prompt_no_rule_no_cont}
\end{figure}

\begin{table}[h!]
\centering
\scriptsize
\caption{Inaccurate Causal Rules used for Input Generation for Generalizability Experiment for \PySIRTEM}
\label{app:tab_inac}
\begin{tabular}{p{0.04\linewidth}|p{0.07\linewidth}|p{0.11\linewidth}|p{0.10\linewidth}|p{0.07\linewidth}|p{0.13\linewidth}|p{0.33\linewidth}}
\toprule
\textbf{Rule ID} & \textbf{Input Window} & \textbf{$X_{t,1}$ Trend} & \textbf{$X_{t,2}$ Trend} & \textbf{Output Window} & \textbf{$Y$ Phase} & \textbf{Symbolic Rule} \\
\hline
R1 & [0, 14] & SharpSpike & Low & [15, 29] & InitialGrowth & 
$\Box_{[0,14]}(\mathrm{SharpSpike}(X_s) \wedge \mathrm{Low}(X_a)) \rightarrow \Diamond_{[15,29]}\mathrm{InitialGrowth}(Y)$ \\
\hline
R2 & [14, 28] & HighStable & HighStable & [29, 43] & PeakFormation & 
$\Box_{[14,28]}(\mathrm{HighStable}(X_s) \wedge \mathrm{HighStable}(X_a)) \rightarrow \Diamond_{[29,43]}\mathrm{PeakFormation}(Y)$ \\
\hline
R3 & [28, 42] & Low & DipThenRise & [43, 57] & PostPeak & 
$\Box_{[28,42]}(\mathrm{Low}(X_s) \wedge \mathrm{DipThenRise}(X_a)) \rightarrow \Diamond_{[43,57]}\mathrm{PostPeak}(Y)$ \\
\hline
R4 & [42, 56] & SharpSpike & HighStable & [57, 71] & Decline & 
$\Box_{[42,56]}(\mathrm{SharpSpike}(X_s) \wedge \mathrm{HighStable}(X_a)) \rightarrow \Diamond_{[57,71]}\mathrm{Decline}(Y)$ \\
\hline
R5 & [56, 70] & Moderate & Low & [71, 85] & Resurgence & 
$\Box_{[56,70]}(\mathrm{Moderate}(X_s) \wedge \mathrm{Low}(X_a)) \rightarrow \Diamond_{[71,85]}\mathrm{Resurgence}(Y)$ \\
\hline
R6 & [70, 100] & HighStable & HighStable & [85, 100] & PeakFormation & 
$\Box_{[70,100]}(\mathrm{HighStable}(X_s) \wedge \mathrm{HighStable}(X_a)) \rightarrow \Diamond_{[85,100]}\mathrm{PeakFormation}(Y)$ \\
\hline
\end{tabular}
\end{table}

\end{document}